\documentclass[11pt]{article}
\usepackage[margin=1in]{geometry}
\usepackage{graphicx}
\usepackage{booktabs}
\usepackage{amsmath}
\usepackage{amssymb}
\usepackage{hyperref}
\usepackage{siunitx}
\usepackage{enumitem}
\usepackage{caption}
\usepackage{microtype}
\usepackage[section]{placeins}

\hypersetup{colorlinks=true, linkcolor=blue, citecolor=blue, urlcolor=blue}
\title{Discovery of a Hematopoietic Manifold in scGPT Yields a Method for Extracting Performant Algorithms from Biological Foundation Model Internals}
\author{Ihor Kendiukhov\textsuperscript{1} \\[4pt]
\textsuperscript{1}Department of Computer Science, University of T\"ubingen, T\"ubingen, Germany \\
\texttt{kendiukhov@gmail.com}}
\date{}

\begin{document}
\maketitle

\begin{abstract}
We report the discovery and extraction of a compact hematopoietic algorithm from the single-cell foundation model scGPT---to our knowledge, the first biologically useful, competitive algorithm extracted from a foundation model via mechanistic interpretability. We show that scGPT internally encodes a compact (\(\sim\!\!8\)--10-dimensional) hematopoietic manifold with significant developmental branch structure, validated on a strict non-overlap Tabula Sapiens external panel (616 anchors, 564,253 cells) and confirmed via frozen-head zero-shot transfer to an independent multi-donor immune panel (trustworthiness \(0.993\), blocked-permutation \(p{=}0.0005\)). To isolate this geometry, we introduce a general three-stage extraction method---direct operator export from frozen attention weights, lightweight learned adaptor, and task-specific readout---that produces a standalone algorithm without target-dataset retraining. In 88-split donor-holdout benchmarks against scVI, Palantir, DPT, CellTypist, PCA, and raw-expression baselines, the extracted algorithm achieves the strongest pseudotime-depth ordering (orientation-independent \(|\rho|{=}0.439\) versus \(0.331\) for the next-best alternative; Wilcoxon BH-\(q \le 2.7{\times}10^{-7}\) on all paired comparisons) and leads on key subtype endpoints (CD4/CD8 AUROC \(0.867\), mono/macro AUROC \(0.951\)). Compared to standard probing of frozen scGPT embeddings with a 3-layer MLP (172k parameters), the extracted head is BH-significantly better on 6/8 classification endpoints while completing a full 12-split evaluation campaign \(34.5\times\) faster (\({\sim}3.4\) versus \({\sim}118\) minutes) with \({\sim}1{,}000\times\) fewer trainable parameters. The exported operator compresses from three pooled attention heads to a single head (L2H5; \(17.5 \rightarrow 5.9\) MB) without statistically significant loss, and further to a rank-64 surrogate (\(0.73\) MB). Mechanistic interpretability of the compact operator reveals a concentrated four-factor core explaining \(66.2\%\) of ablation impact, with factors resolving into explicit T/lymphoid, B/plasma, granulocytic, and monocyte/macrophage gene programs. A supplementary second-manifold validation (intercellular communication geometry) confirms that the extraction method generalizes beyond hematopoiesis.
\end{abstract}

\section{Introduction}
Foundation models for biology are increasingly powerful but largely opaque. Built on the transformer architecture \cite{vaswani2017attention}, single-cell foundation models such as scGPT \cite{zheng2024scgpt}, Geneformer \cite{theodoris2023geneformer}, scBERT \cite{yang2022scbert}, scFoundation \cite{hao2024scfoundation}, and UCE \cite{rosen2023universal} build on earlier latent-variable approaches like scVI \cite{lopez2018scvi,gayoso2021totalvi} to learn rich representations of cellular state. Yet the question of \emph{what biological knowledge these models actually encode}---and whether it can be extracted and reused---remains largely open, motivating the application of mechanistic interpretability methods \cite{olah2020zoom,elhage2021mathematical} to biological foundation models. Mechanistic interpretability has made rapid progress in language models, revealing superposition phenomena \cite{elhage2022superposition}, complete circuits for specific behaviors \cite{wang2023interpretability,neel2023progress}, methods for automated neuron interpretation \cite{bills2023language}, and evidence for linear geometric structure in learned representations \cite{park2024linear}. In biological models specifically, attention analysis of protein language models has shown that transformer heads can encode meaningful biochemical relationships \cite{vig2021bertology,clark2019bert}. Recent work on single-cell foundation models has begun probing what these models learn: systematic evaluation of attention patterns in scGPT and Geneformer revealed that attention captures gene co-expression structure rather than unique regulatory signal, with simple gene-level baselines dominating pairwise attention edges for perturbation prediction \cite{kendiukhov2026attention}. Sparse autoencoder decomposition of both models uncovered over 107,000 interpretable features organized into biological pathways, yet only 6.2\% of transcription factors showed regulatory-target-specific responses, indicating minimal causal regulatory logic \cite{kendiukhov2026sparse}. Causal circuit tracing further revealed that both architectures converge on inhibitory-dominant computational motifs with approximately 53\% biological coherence, suggesting shared representational principles across model families \cite{kendiukhov2026causal}. However, the question of whether the structured biological knowledge these models \emph{do} encode can be extracted as a reusable algorithm has not been addressed.

A key precedent comes from protein language models. Protein structure prediction \cite{jumper2021alphafold} and evolutionary sequence models \cite{rives2021protein,lin2023esm,nguyen2024evo2} have demonstrated that large-scale pretraining captures deep biological structure. The Goodfire phylogeny-manifold program showed that Evo2 \cite{nguyen2024evo2} internally encodes a compact manifold corresponding to the evolutionary tree of life, recoverable through rigorous geometric auditing with external biological rulers \cite{goodfire2025,rives2021protein,lin2023esm}. That result established that biologically meaningful manifolds can be discovered inside foundation models. More broadly, methods for localizing knowledge in specific transformer components \cite{meng2022locating,geva2021transformer} and for discovering causal structure through intervention \cite{geiger2021causal} provide a methodological foundation for the extraction approach we pursue here.

We ask: can we go further? Specifically, can we (i) discover analogous biological manifolds in single-cell foundation models, (ii) extract them as reusable algorithms, and (iii) compress them into compact, interpretable operators? This paper answers all three questions affirmatively for hematopoiesis in scGPT, yielding what is, to our knowledge, the first competitive algorithm extracted from a foundation model via mechanistic interpretability.

Our approach connects to the rich tradition of geometric manifold learning \cite{isomap2000,roweis2000nonlinear,coifman2006diffusion,umap2018,moon2019phate} and trajectory inference in single-cell biology \cite{trapnell2014monocle,qiu2017reversed,haghverdi2016dpt,street2018slingshot,wolf2019paga,saelens2019comparison}, including RNA velocity approaches \cite{lamanno2018rna,bergen2020scvelo}, single-cell data integration methods \cite{stuart2019comprehensive,korsunsky2019harmony,luecken2022benchmarking}, and reference-atlas transfer learning \cite{lotfollahi2022scarches}. However, rather than fitting manifolds to observed gene expression, we extract geometric structure from the \emph{internal} attention operators of a pretrained model and evaluate it under strict holdout, confound, and external-transfer controls. For biological grounding, we use Tabula Sapiens \cite{tabulasapiens2022}, part of the broader Human Cell Atlas effort \cite{regev2017humancellatlas}, and curated regulatory resources \cite{dorothea2019,trrust2018}. The discovery was conducted through a two-phase autonomous research loop: a broad hypothesis search that screened dozens of candidates against pre-registered quality gates, followed by focused investigation of the surviving positive branch (Section~\ref{sec:autoloop}).

\paragraph{Main contributions.}
\begin{enumerate}[leftmargin=1.5em]
    \item \textbf{Manifold discovery.} We discover a compact hematopoietic manifold in scGPT, validated internally (trustworthiness \(0.993\), blocked-permutation \(p{=}0.0005\)) and externally on a strict non-overlap Tabula Sapiens panel (616 anchors, 564,253 cells) and a separate multi-donor panel in frozen zero-shot mode. 
    \item \textbf{Extraction method.} We introduce a three-stage pipeline---direct operator export, lightweight learned adaptor, task-specific readout---that isolates transferable biological geometry from frozen model weights without target-dataset retraining.
    \item \textbf{Competitive extracted algorithm.} The extracted algorithm significantly outperforms scVI, Palantir, DPT, CellTypist, PCA, and raw-expression baselines on pseudotime-depth alignment (88 donor-holdout splits; all paired BH-\(q \le 2.7{\times}10^{-7}\)) and leads on key subtype classification endpoints.
    \item \textbf{Multi-stage compaction.} The exported operator compresses from pooled attention heads to a single head (L2H5; \(17.5 \rightarrow 5.9\) MB) without significant loss, and further to a rank-64 surrogate (\(0.73\) MB).
    \item \textbf{Mechanistic interpretability.} Factor ablation and sparse factorization of the compact operator reveal a four-factor core (\(66.2\%\) of ablation impact) resolving into explicit hematopoietic gene programs, providing among the most detailed mechanistic decompositions of a foundation-model-derived algorithm.
\end{enumerate}

\section{Methods}

\subsection{Autonomous research loop}
\label{sec:autoloop}
The study was conducted through a two-phase autonomous research loop driven by an AI executor--reviewer pair operating under pre-registered quality gates, extending the emerging paradigm of AI-driven scientific discovery \cite{lu2024aiscientist}.

\textbf{Phase 1: Broad hypothesis search.} The loop explored a large space of candidate manifold hypotheses by systematically varying biological targets (developmental ordering, regulatory structure, communication geometry), featurization strategies (attention drift, raw embeddings, mixed operators), and geometric fitting methods (Isomap, geodesic-MDS, LET) across the scGPT attention tensor. Each candidate was evaluated against fixed quantitative gates---trustworthiness \(\ge 0.80\), holdout correlation \(\ge 0.20\) on random/donor/clade splits, and blocked-permutation \(p \le 0.001\)---with an adversarial reviewer role rejecting candidates that failed any gate or showed confound sensitivity. This phase operated at scale: dozens of hypothesis branches were instantiated, tested, and rejected before a positive signal emerged. The key output of Phase~1 was the identification of hypothesis H65 (hematopoietic developmental manifold) as the first branch to pass all gates with a robust positive signal, while a paired null branch (H66, same pipeline with a different biological target) failed, confirming that the signal was branch-specific rather than a pipeline artifact.

\textbf{Phase 2: Focused investigation.} Once the autonomous loop identified H65 as a robust positive branch, the research transitioned to manual, author-led investigation. This phase included: methodological closure tests (objective ablations, confidence intervals, structured holdouts), strict non-overlap external transfer validation, extraction and benchmarking of the manifold as a standalone algorithm, multi-stage operator compaction, and mechanistic interpretability via factor ablation and sparse factorization. While the same quantitative gate logic from Phase~1 was retained as a quality standard, the experimental design, analysis, and interpretation in Phase~2 were conducted by the authors rather than by the autonomous loop. A second branch (H38, intercellular communication geometry) was also carried through both phases as an independent generalization test (Supplementary~S32--S37). Full iteration logs and the Phase~1 search trajectory are documented in Supplementary~S1.

\subsection{Data and anchor construction}
We derived donor\(\times\)tissue\(\times\)cell-type \textbf{anchors}---aggregate representations (centroids) computed by averaging scGPT embeddings over all cells sharing a donor, tissue, and cell-type label---from Tabula Sapiens \cite{tabulasapiens2022} immune data processed through the frozen scGPT whole-human checkpoint \cite{zheng2024scgpt}. The \textbf{internal panel} used 298 anchors across 7 hematopoietic branch groups (5 tissues). The \textbf{strict non-overlap external panel} was built by excluding all internal observation IDs, mapping remaining cells to the same stage ontology, and aggregating: 564,253 cells, 616 anchors, 21 donors, 27 tissues, 34 fine stages across 7 branches. A separate \textbf{multi-donor immune panel} (7 cohorts, 165 anchors, 10,819 sampled cells) was reserved for frozen zero-shot transfer confirmation (protocol in Supplementary~S4). No external data were used during any training step.

\subsection{Three-stage extraction pipeline}
\label{sec:extraction_pipeline}
We treat the manifold as an extracted algorithm rather than a visualization artifact:
\[
x \xrightarrow{\text{Stage 1: frozen operator}} f(x) \xrightarrow{\text{Stage 2: learned adaptor }g_{\theta}} z \xrightarrow{\text{Stage 3: task readout }h_{\phi}} \hat{y}.
\]

\textbf{Stage 1: Direct operator extraction.} From the frozen scGPT checkpoint, we read native attention operators \(A_{\ell,h}\in\mathbb{R}^{1200\times 1200}\) (the value-projection weight matrices from layer~\(\ell\), head~\(h\), which define how each gene's embedding is linearly transformed within that head) and construct a fixed feature map. The main benchmark uses a pooled drift operator:
\[
f_{\mathrm{drift}}(x)=\big[(xA_{\mathrm{early}}-xA_{\mathrm{mid}})\ ;\ (xA_{\mathrm{mid}}-xA_{\mathrm{late}})\big]\in\mathbb{R}^{2400},
\]
where \(A_{\mathrm{early}},A_{\mathrm{mid}},A_{\mathrm{late}}\) average native heads over early, middle, and late layer blocks. The intuition is that successive transformer layers progressively refine gene representations; the pairwise differences capture how each cell's representation \emph{changes} across layers, and this representational drift may encode developmental trajectory information. No target labels are used; no parameters are optimized.

\textbf{Stage 2: Lightweight learned adaptor.} A small head \(g_{\theta}\) is trained on internal data only, mapping fixed features to a task-agnostic manifold latent \(z\) (\(d=10\)). We use a Latent Embedding Transfer (LET) objective:
\begin{align}
z &= W_{\mathrm{enc}}(x-b), \quad
\hat{d}_{ij} = \beta\,\arccos\!\left(\cos(z_i,z_j)\right), \\
\mathcal{L} &= \lVert\hat{d}-d_{\mathrm{target}}\rVert^2 + \alpha\lVert W_{\mathrm{enc}}^{+} z + b - x\rVert^2.
\end{align}
Here \(d_{\mathrm{target}}\) are pairwise biological distances derived from a curated hematopoietic stage ontology (e.g., the number of developmental steps separating two cell types), so the first term encourages the latent geometry to reflect developmental proximity, and the second term is a reconstruction regularizer. Quality gates require trustworthiness (a measure of local neighborhood preservation: the fraction of each point's nearest neighbors in the high-dimensional input that remain neighbors in the low-dimensional embedding \cite{venna2006trustworthiness}) \(\ge 0.80\) and random/donor/clade holdout correlations \(\ge 0.20\) (full equation-level details in Supplementary~S3). We trained three head variants: an anchor-trained head, a cell-trained head (6,925 cells, stage-balanced), and a hybrid head with topology preservation (Supplementary~S5--S6). An artifact-closure protocol ensured that all claims survive structured holdout and permutation controls (Supplementary~S2).

\textbf{Stage 3: Task-specific readout.} Small probes \(h_{\phi}\) are trained on top of \(z\) for classification or pseudotime regression. These are not part of the shared representation.

\subsection{Head/layer attribution and multi-stage compaction}
Drawing on methods for localizing factual knowledge in specific model components \cite{meng2022locating,geva2021transformer}, we scanned all 96 units of the \(12\times 8\) scGPT attention tensor. Each unit \((\ell,h)\) was evaluated via single-head feature map \(x \mapsto x A_{\ell,h}\) with LET-10D trained on internal anchors and tested zero-shot on external anchors. Compact operators replace the pooled drift map with a weighted sum of top-ranked heads:
\[
\widetilde{A}_{k}=\sum_{i=1}^{k}\alpha_i A_{\ell_i,h_i}.
\]
Further compression uses truncated-SVD surrogates \(\widetilde{A}_{k}\approx U_rV_r^{\top}\) (\(r\in\{8,16,32,64\}\)) and hard sparse pruning (top-\(k\) read and write genes per retained factor, where ``read genes'' are those with highest input loadings and ``write genes'' are those with highest output loadings in each rank-1 factor of the SVD). Each compression stage retrains only the lightweight adaptor and task probes. Factor necessity was assessed via fixed-probe leave-one-out ablation. Terminology, compression stages, and the full operator hierarchy are detailed in Supplementary~S12.

\subsection{Benchmark evaluation protocol}
We benchmarked extracted heads against raw-expression (1200 \(\log(1{+}x)\) genes), PCA-10, SVD-10, CellTypist \cite{dominguez2022celltypist}, scVI-10D \cite{lopez2018scvi}, Scanpy-DPT \cite{wolf2018scanpy,haghverdi2016dpt}, and Palantir \cite{setty2019palantir} on the strict non-overlap external panel. The finalized Robust-V2 campaign used 88 grouped donor-holdout splits (100,000 training cells per split). Metrics: branch/stage balanced accuracy and macro-F1, CD4/CD8 and mono/macro AUROC, and pseudotime-depth Spearman correlation (the rank correlation between the inferred pseudotime ordering and the known ontology-based developmental stage depth). All pseudotime was evaluated with donor-local test-only protocol. Paired statistical comparisons used split-level Wilcoxon signed-rank tests with Benjamini--Hochberg false discovery rate (BH-FDR) correction \cite{benjamini1995controlling}. Statistical controls and robustness tests (confidence intervals, structured holdout correlations) are detailed in Supplementary~S7; full benchmark protocol details in Supplementary~S8. Additional ablations (embedding baselines, probe-depth variants, direct-MLP learnability) are reported in Supplementary~S19--S22.

\section{Results}

\subsection{Discovery of a hematopoietic manifold in scGPT}
\begin{figure}[t]
    \centering
    \includegraphics[width=0.96\linewidth]{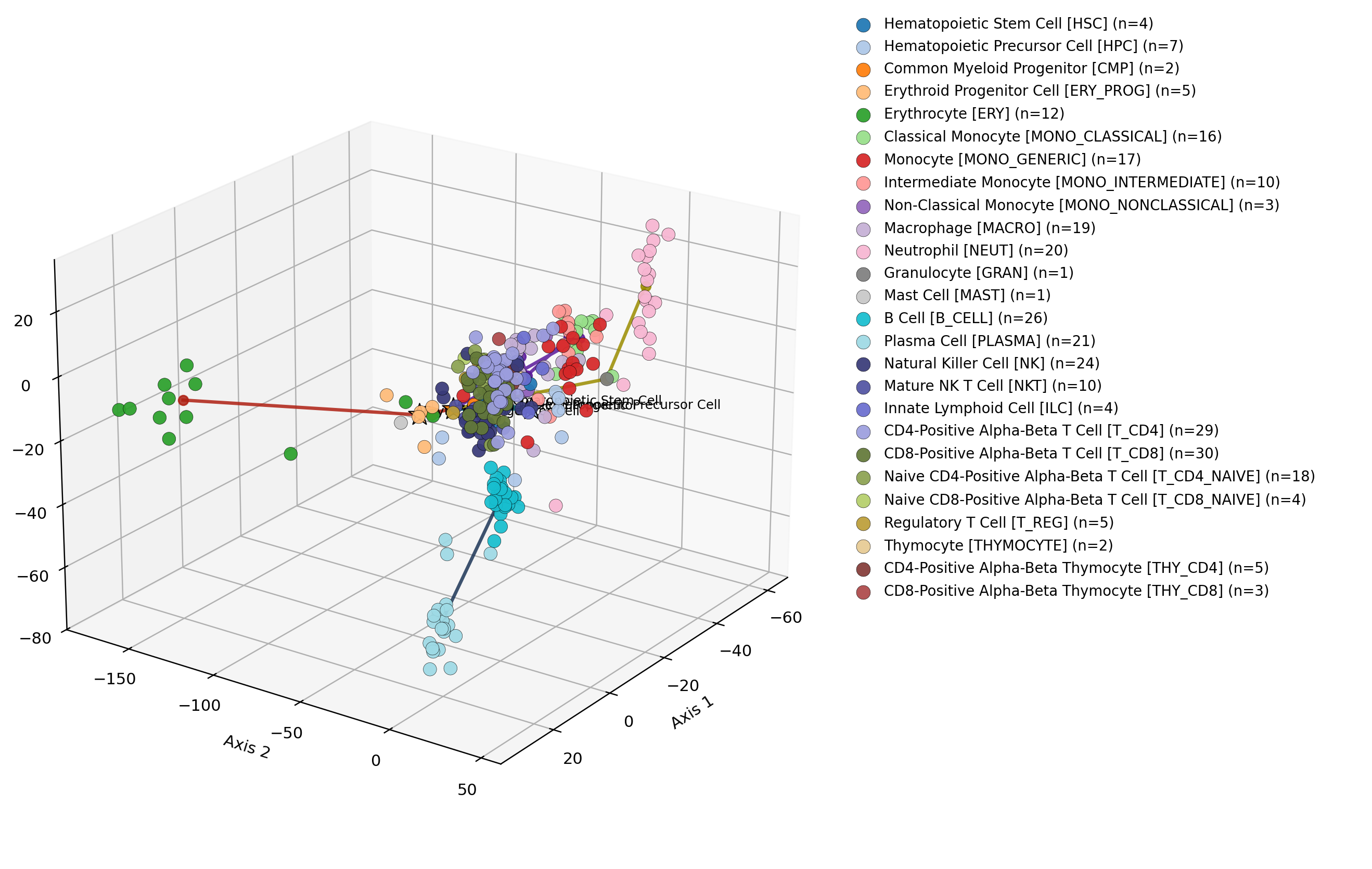}
    \caption{Hematopoietic manifold discovered in scGPT attention geometry (internal panel). Distinct branch structure emerges around stem/progenitor, erythroid, granulocytic, monocyte/macrophage, and lymphoid/T-cell regions.}
    \label{fig:expanded3d}
\end{figure}

The Phase~1 broad search (Section~\ref{sec:autoloop}) converged on hypothesis H65 as the robust positive branch after rejecting dozens of alternative candidates. On the internal panel, H65 achieved high manifold fidelity (trustworthiness \(0.993\), geodesic--biological correlation \(0.835\), blocked-permutation \(p{=}0.0005\)), while a paired null branch (H66) failed gates. The manifold exhibits distinct hematopoietic branch structure (Figure~\ref{fig:expanded3d}) with significant developmental ordering along erythroid (\(\rho{=}0.768\), \(p{=}0.0017\)), granulocytic (\(\rho{=}0.568\), \(p{=}0.0033\)), and trunk HSC/HPC/CMP (hematopoietic stem cell / hematopoietic progenitor cell / common myeloid progenitor; \(\rho{=}0.611\), \(p{=}0.018\)) trajectories (Supplementary~S13). The recovered branch structure aligns with the established hierarchical model of hematopoiesis \cite{orkin2008hematopoiesis,laurenti2018haematopoietic}, including the continuous lineage commitment observed in progenitor populations \cite{velten2017human,paul2015transcriptional} and the direct link between transcriptional state and cell fate established by clonal barcoding \cite{weinreb2020lineage}. The effective dimensionality is \(\sim\!\!8\)--10: external transfer objective and trustworthiness both improve from \(d{=}3\) through \(d{=}10\) (Supplementary~S14).

\subsection{External validation and zero-shot transfer}
A small independent lung panel showed strong global association but failed robust transfer gates (low trustworthiness, weak clade transfer). In contrast, the strict non-overlap Tabula Sapiens panel passed all gates under the same LET protocol (Table~\ref{tab:transfer}). A separate frozen-head zero-shot test on an independent multi-donor immune panel (7 cohorts, 165 anchors) further confirmed transferability without any retraining (Table~\ref{tab:zeroshot_external}).

\begin{table}[t]
\centering
\caption{LET holdout robustness under shared gates. Trust.\ = trustworthiness; Random/Donor/Branch = Spearman correlation between predicted and target distances on random, donor-held-out, and branch-held-out test splits, respectively.}
\label{tab:transfer}
\small
\begin{tabular}{lcccc}
\toprule
Panel & Trust. & Random & Donor & Branch \\
\midrule
Internal (H65, \(d{=}10\)) & 0.979 & 0.495 & 0.496 & 0.382 \\
External lung (\(d{=}6\)) & 0.617 & 0.236 & 0.687 & 0.058 \\
External Tabula non-overlap (\(d{=}10\)) & 0.985 & 0.594 & 0.596 & 0.453 \\
\bottomrule
\end{tabular}
\end{table}

\begin{table}[t]
\centering
\caption{Frozen-head zero-shot transfer on a separate external multi-donor panel.}
\label{tab:zeroshot_external}
\small
\resizebox{\linewidth}{!}{
\begin{tabular}{lccccccc}
\toprule
Panel & Anchors & Donors & Stages & Corr & Corr\(_{\mathrm{resid}}\) & Trustworthiness & Blocked \(p\) \\
\midrule
External multi-donor zero-shot & 165 & 7 & 7 & 0.500 & 0.476 & 0.993 & 0.0005 \\
\bottomrule
\end{tabular}
}
\end{table}

\begin{figure}[t]
    \centering
    \includegraphics[width=0.9\linewidth]{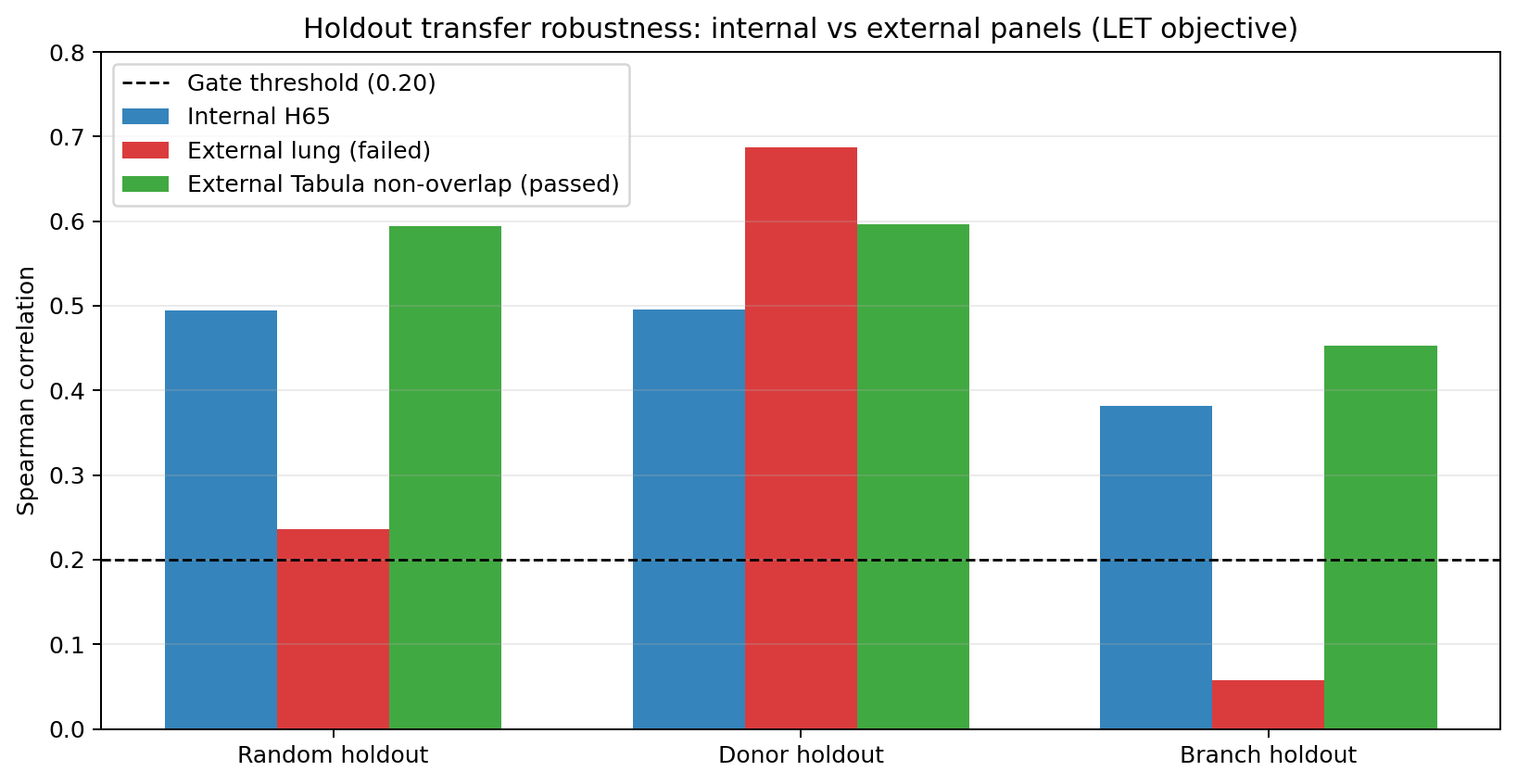}
    \caption{Holdout transfer comparison. The strict non-overlap external panel passes all gates, while the lung panel fails robust transfer despite strong global association.}
    \label{fig:transfer_bars}
\end{figure}

We trained three head variants via Stage~2 of the extraction pipeline: an anchor-trained head, a cell-trained head (6,925 cells), and a hybrid head with topology preservation. The cell-trained head improved subtype separability (CD4/CD8 AUC \(0.744{\rightarrow}0.886\), mono/macro AUC \(0.759{\rightarrow}0.937\)) at a small cost in branch compactness; the hybrid head offered a Pareto tradeoff favoring branch readability (Supplementary~S15--S16).

\subsection{The extracted algorithm is competitive}
We evaluated the extracted manifold head as a standalone algorithm against established methods. Table~\ref{tab:posthoc_largecell_iter1} reports the finalized Robust-V2 benchmark (88 grouped donor-holdout splits, 100k training cells per split, donor-local test-only pseudotime).

\begin{table}[t]
\centering
\caption{Large-cell Robust-V2 pooled donor-holdout benchmark on strict non-overlap mapped cells (100k training cells per split). Values are pooled donor-level means across 88 grouped splits. Coverage varies by method/metric due to valid-run filtering (pseudotime: 88 splits for most methods, 70 for Palantir, 36 for scVI).}
\label{tab:posthoc_largecell_iter1}
\small
\resizebox{\linewidth}{!}{
\begin{tabular}{lccccccc}
\toprule
Method & Branch bal. acc. & Branch macro-F1 & Stage bal. acc. & Stage macro-F1 & CD4/CD8 AUROC & Mono/Macro AUROC & Pseudotime Spearman \\
\midrule
H65 anchor head (10D) & 0.750 & 0.557 & 0.465 & 0.184 & 0.803 & 0.934 & 0.093 \\
H65 cell head (10D) & 0.828 & 0.584 & 0.552 & 0.216 & 0.867 & 0.951 & 0.249 \\
H65 hybrid conservative (10D) & 0.815 & 0.599 & 0.542 & 0.217 & 0.858 & 0.949 & 0.132 \\
Raw baseline (1200 log1p) & 0.826 & 0.595 & 0.567 & 0.219 & 0.805 & 0.833 & -0.044 \\
PCA-10 baseline & 0.733 & 0.543 & 0.503 & 0.194 & 0.842 & 0.944 & -0.007 \\
SVD-10 baseline & 0.731 & 0.541 & 0.499 & 0.192 & 0.842 & 0.936 & 0.002 \\
CellTypist stage-probability baseline & 0.703 & 0.588 & 0.393 & 0.226 & 0.785 & 0.855 & -0.064 \\
scVI-10D baseline & 0.793 & 0.551 & 0.512 & 0.205 & 0.849 & 0.916 & -0.103 \\
Scanpy-DPT baseline & 0.750 & 0.606 & 0.451 & 0.235 & 0.793 & 0.947 & -0.089 \\
Palantir baseline & 0.759 & 0.621 & 0.423 & 0.242 & 0.750 & 0.946 & -0.049 \\
\bottomrule
\end{tabular}
}
\end{table}

Classification remains mixed by endpoint. Stage/branch macro-F1 is highest for diffusion-style baselines (Palantir and Scanpy-DPT), while raw-expression is highest on stage balanced accuracy. The extracted cell head remains strongest on branch balanced accuracy, CD4/CD8 AUROC, and mono/macro AUROC.

\begin{figure}[t]
    \centering
    \includegraphics[width=0.95\linewidth]{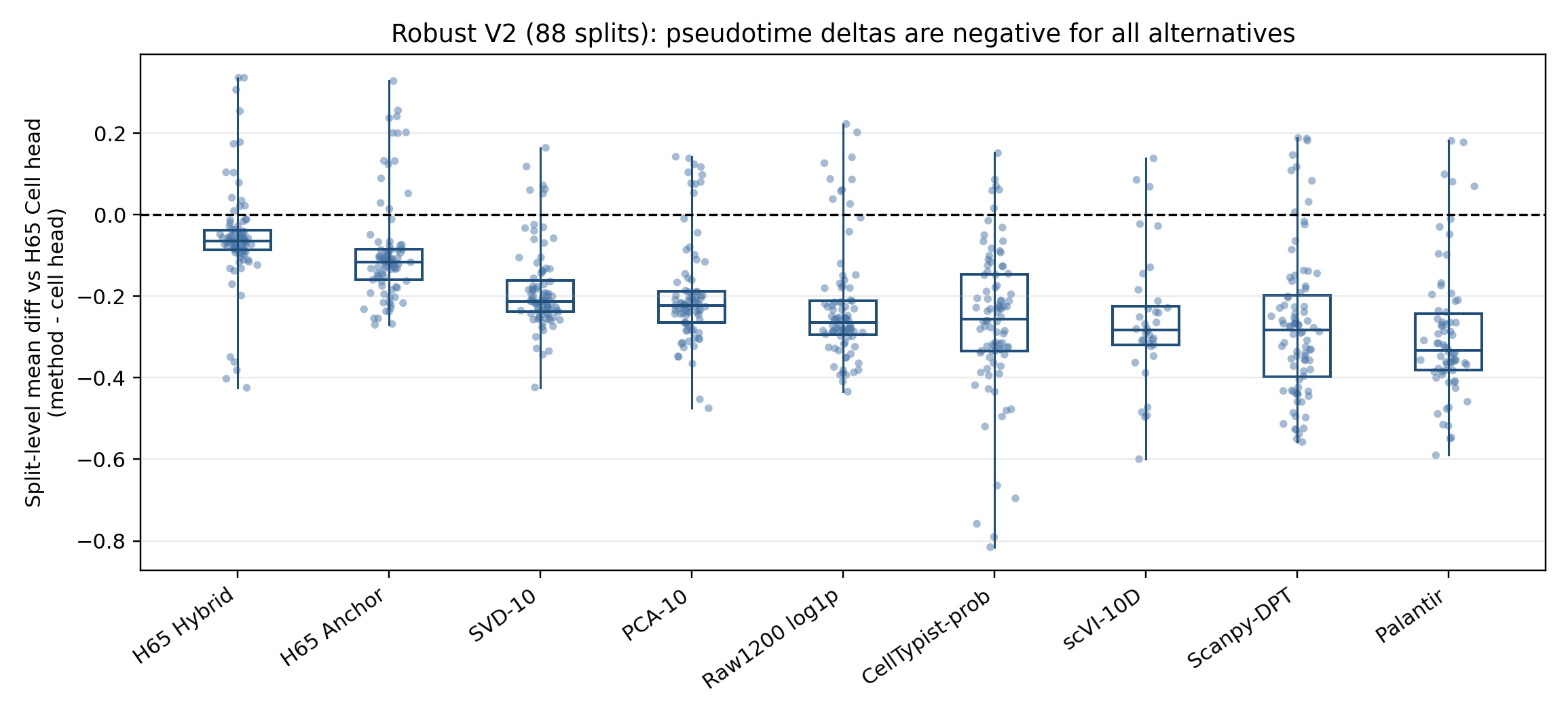}
    \caption{Robust-V2 split-level pseudotime deltas versus H65 cell head (method minus cell head). All alternatives are below zero.}
    \label{fig:robustv2_pt_deltas}
\end{figure}

For pseudotime-depth ordering, the extracted cell head leads all compared methods. Because pseudotime orientation is arbitrary across donor-local evaluations (each donor's pseudotime is computed independently, and the algorithm may assign increasing values toward either stem or mature cells), we report the orientation-independent \(|\rho|\) (see Section~\ref{sec:s17} for the full orientation analysis): the cell head achieves \(|\rho|{=}0.439\) versus \(0.364\) (hybrid), \(0.331\) (Palantir), \(0.279\) (SVD), \(0.274\) (scVI), \(0.266\) (raw), \(0.261\) (DPT), \(0.259\) (PCA), \(0.253\) (CellTypist), and \(0.235\) (anchor). In paired split-level tests versus the cell head, all method-minus-head deltas are negative with Wilcoxon BH-\(q\le2.7\times10^{-7}\) in the 88-split pooled analysis (Table~\ref{tab:posthoc_pairwise_pt_v2} and Figure~\ref{fig:robustv2_pt_deltas}).

\begin{table}[t]
\centering
\caption{Robust-V2 paired pseudotime statistics versus H65 cell head (split-level mean differences; method minus cell head).}
\label{tab:posthoc_pairwise_pt_v2}
\small
\begin{tabular}{lccc}
\toprule
Method & \(n_{\mathrm{splits}}\) & Mean diff & Wilcoxon BH-\(q\) \\
\midrule
H65 hybrid conservative (10D) & 88 & -0.056 & \(2.7\times10^{-7}\) \\
H65 anchor head (10D) & 88 & -0.092 & \(2.7\times10^{-7}\) \\
SVD-10 baseline & 88 & -0.185 & \(9.7\times10^{-15}\) \\
PCA-10 baseline & 88 & -0.194 & \(2.6\times10^{-14}\) \\
Raw baseline (1200 log1p) & 88 & -0.226 & \(2.5\times10^{-14}\) \\
CellTypist stage-probability baseline & 88 & -0.260 & \(7.0\times10^{-15}\) \\
scVI-10D baseline & 36 & -0.261 & \(4.1\times10^{-9}\) \\
Scanpy-DPT baseline & 88 & -0.272 & \(2.6\times10^{-14}\) \\
Palantir baseline & 70 & -0.293 & \(3.7\times10^{-12}\) \\
\bottomrule
\end{tabular}
\end{table}


The extraction gain is not reducible to trivially fitting a small MLP on raw expression. A direct raw-expression MLP (1200$\rightarrow$64) learns substantial stage signal (stage balanced accuracy 0.499--0.504 versus 0.468 for the extracted head) but fails to match the extracted head's subtype-sensitive advantages: CD4/CD8 AUROC 0.828 versus 0.856, mono/macro AUROC 0.920--0.926 versus 0.956 (all BH-significant; Supplementary~S22). The scGPT-derived geometric structure provides subtype separation that raw-expression probes cannot recover.

A stronger test asks whether the advantage persists when frozen scGPT embeddings are paired with a deeper non-linear readout. In 12-split repeated evaluation, a frozen scGPT avg-pool embedding with a 3-layer MLP (172k--175k trainable parameters) is BH-significantly worse than the extracted 10D head on 6/8 endpoints, including branch macro-F1 ($\Delta{=}{-}0.045$, BH-$q{=}0.010$) and stage macro-F1 ($\Delta{=}{-}0.078$, BH-$q{=}0.0013$); the deeper MLP matches or closes the gap only on CD4/CD8 and mono/macro AUROC (Supplementary~S19--S20). When reducing to a comparable 4D head, the extracted operator wins on 5/8 endpoints with BH correction. A probe-depth sweep on the extracted head itself (Supplementary~S21) shows that a 2-layer MLP readout yields BH-significant gains on CD4/CD8 AUROC ($+0.0045$) but worsens stage endpoints, confirming that the linear default is a robust Pareto choice.

Beyond accuracy, the extracted head is substantially faster. One-time preparation (drift operator computation and LET head training) takes ${\sim}100$~seconds versus ${\sim}5{,}164$~seconds for running all cells through the frozen scGPT encoder to obtain embeddings---a $52\times$ gap. Per-split training costs ${\sim}9$~seconds (2~s probe training + 7~s drift-feature inference) versus ${\sim}159$~seconds for the 3-layer MLP ($18\times$), and the full 12-split campaign completes in ${\sim}205$~seconds versus ${\sim}7{,}074$~seconds ($34.5\times$ faster). The extracted head requires only 5--170 trainable parameters per downstream task, versus 172k--175k for the frozen-embedding MLP path (Supplementary~S23).

Table~\ref{tab:method_comparison_summary} consolidates these comparisons across performance, training cost, and deployment footprint. The frozen scGPT path additionally requires running all ${\sim}140$k cells through the full 12-layer transformer to obtain embeddings (${\sim}86$~minutes), making it the most expensive preparation step by far. The direct raw-expression MLP avoids scGPT entirely but requires ${\sim}120$--$188$~s per split due to its 1,200-dimensional input, and cannot recover the subtype geometry that scGPT's internal representations provide.

\begin{table}[t]
\centering
\caption{Summary comparison of the three evaluated approaches on the 12-split repeated benchmark. Performance values are pooled donor-level means. Training times from Supplementary~S23; parameter counts from model architectures. BH-sig.\ wins/losses count endpoints (out of 8) on which the method is BH-significantly better/worse than the extracted head.}
\label{tab:method_comparison_summary}
\small
\resizebox{\linewidth}{!}{
\begin{tabular}{lccc}
\toprule
 & Extracted 10D head + MLP2 & Frozen scGPT + 3-layer MLP & Direct raw MLP (\(128{\rightarrow}64\)) \\
\midrule
\multicolumn{4}{l}{\emph{Key classification metrics}} \\
\quad Branch bal.\ acc. & \textbf{0.827} & 0.810 & 0.818 \\
\quad Stage bal.\ acc. & 0.468 & 0.403 & \textbf{0.504} \\
\quad CD4/CD8 AUROC & \textbf{0.856} & 0.817 & 0.827 \\
\quad Mono/Macro AUROC & \textbf{0.956} & 0.946 & 0.920 \\
\midrule
\multicolumn{4}{l}{\emph{Training and deployment cost}} \\
\quad One-time preparation & 100 s (drift + head) & 5{,}164 s (encoder fwd pass) & 172 s (log-normalize) \\
\quad Per-split training & 2 s (probe) + 7 s (inference) & 159 s (MLP training) & 120--188 s (MLP training) \\
\quad Full 12-split campaign & ${\sim}205$ s (${\sim}3.4$ min) & ${\sim}7{,}074$ s (${\sim}118$ min) & ${\sim}1{,}600$--$2{,}400$ s \\
\quad Trainable parameters / task & 5--170 & 172{,}610--174{,}690 & 77{,}000--164{,}000 \\
\quad Deployable artifact & 5.9 MB (compact head) & ${\sim}400$ MB (frozen encoder) & $<$1 MB \\
\midrule
BH-sig.\ wins vs extracted head & --- & 0 / 8 & 2 / 8 (stage only) \\
BH-sig.\ losses vs extracted head & --- & 6 / 8 & 3 / 8 (subtype) \\
\bottomrule
\end{tabular}
}
\end{table}

Unsupervised branch recovery, by contrast, is a weakness. When the extracted head's output is fed through standard unsupervised wrappers (Scanpy Leiden+PAGA or Palantir fate assignment), raw expression performs comparably or better on partition metrics such as the Adjusted Rand Index (ARI). The extracted head's advantage is concentrated in supervised classification and pseudotime ordering, not in unsupervised topology recovery (Supplementary~S18).


\subsection{Multi-stage compaction}
To localize the biological signal within scGPT, we scanned all 96 attention heads (12 layers \(\times\) 8 heads). A single head---Layer~2, Head~5 (L2H5)---carried the strongest transferable developmental geometry (residualized correlation \(0.634\), i.e., the correlation between predicted and target distances after regressing out donor and tissue confounds; trustworthiness \(0.985\)), competitive with the three-head pooled variants (pooled-anchor \(0.605\), pooled-cell \(0.612\), hybrid \(0.632\); Supplementary~S29). Pooled heads retain an advantage on subtype margins (CD4/CD8 AUROC \(0.911\) hybrid versus \(0.711\) L2H5 in 3D), motivating their use in the benchmark, but the single-head result shows the signal is surprisingly localized. The compact operator \(\widetilde{A}_{1}=\alpha_1 A_{2,5}\) preserves classification utility without BH-significant loss on all eight repeated endpoints, while reducing the deployable artifact from \({\sim}17.5\) MB (three pooled heads) to \({\sim}5.9\) MB (Table~\ref{tab:compact_summary}).

Further compression via truncated-SVD yields viable surrogates down to rank 64 (\(0.73\) MB), though with BH-significant losses on 5/8 endpoints versus the dense compact operator. More aggressive ranks (32, 16, 8) degrade substantially. Hard sparse pruning (16 factors, 60 read/write genes each; 124 KB deployable package) yields an interpretable but baseline-level-accuracy surrogate.

\begin{table}[t]
\centering
\caption{Compaction chain: from pooled drift operator to single-head and low-rank surrogates. Losses are relative to the full extracted 10D head under paired BH-corrected tests.}
\label{tab:compact_summary}
\small
\begin{tabular}{lccc}
\toprule
Operator variant & Size & BH-sig.\ losses (of 8) & Key note \\
\midrule
Full drift (3 pooled heads) & 17.5 MB & --- & Reference \\
Compact top1 (L2H5) & 5.9 MB & 0/8 & Best single head \\
Top1 + rank 64 & 0.73 MB & 5/8 & Viable low-rank boundary \\
Top1 + rank 32 & 0.37 MB & 6/8 & Degraded \\
Hard sparse (16 factors, 60 genes) & 0.12 MB & 7/8 & Interpretable, baseline-level \\
\bottomrule
\end{tabular}
\end{table}




\subsection{Mechanistic interpretability of the compact operator}
\begin{figure}[!htbp]
    \centering
    \includegraphics[width=\linewidth]{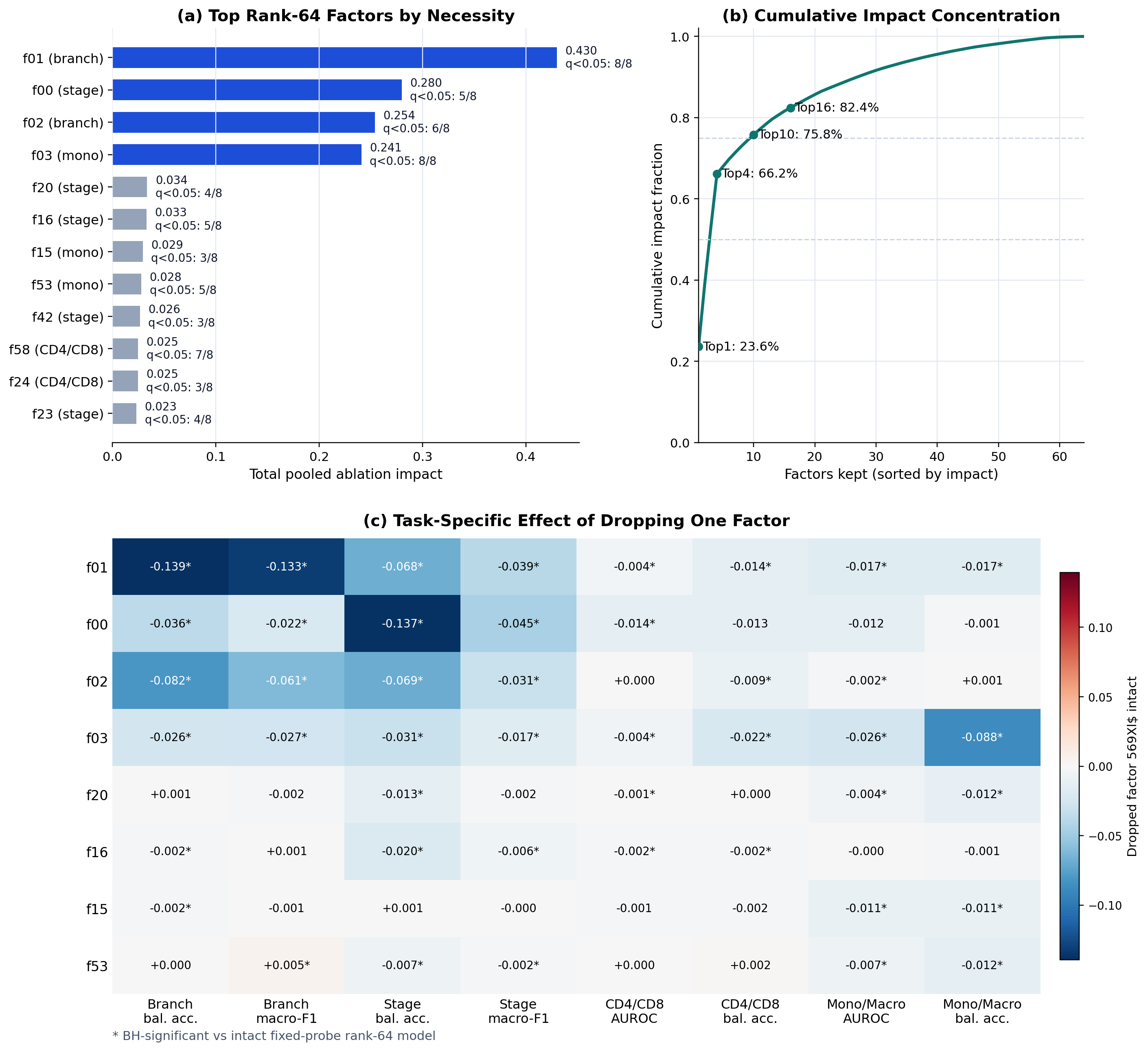}
    \caption{Fixed-probe leave-one-factor-out ablation for the rank-64 compact operator. (a)~Total pooled ablation impact for the 12 most impactful factors. (b)~Cumulative concentration curve---the top four factors explain \(66.2\%\) of total ablation impact. (c)~Per-task drop magnitudes; asterisks denote BH-significant loss versus the intact model.}
    \label{fig:factor_ablation}
\end{figure}

The rank-64 low-rank factorization (via truncated SVD, yielding 64 rank-1 components, each defined by a pair of input and output gene-loading vectors) enables direct mechanistic analysis. Fixed-probe leave-one-factor-out ablation reveals highly concentrated necessity: the top four factors (f01, f02, f00, f03) explain \(66.2\%\) of total pooled ablation impact (the summed classification accuracy drop across all endpoints when a factor is removed) (Figure~\ref{fig:factor_ablation}). A core-sufficiency test shows that this four-factor subset is necessary but not sufficient: branch balanced accuracy drops from \(0.820\) (intact rank-64) to \(0.572\) (four-factor core), CD4/CD8 AUROC from \(0.846\) to \(0.699\), and all eight endpoints are BH-significantly worse than the intact model (Supplementary~S30). An exhaustive 15-subset interaction sweep over the four core factors reveals task-specialized circuitry rather than a monotone additive ladder: branch/stage classification is best served by the full four-factor set, while mono/macro separation peaks at the pair \{f01, f03\} (AUROC \(0.902\), \(94.9\%\) of intact) and CD4/CD8 at a triple centered on f00+f03 (AUROC \(0.703\), \(83.1\%\) of intact).

Cached local marker enrichment and external pathway analysis using Gene Ontology annotations \cite{ashburner2000gene} resolve these core factors into explicit biological programs (Table~\ref{tab:gene_programs}):
\begin{itemize}[leftmargin=1.5em]
    \item \textbf{f01}: branch-routing factor, dominant on branch classification; monocyte/macrophage-aligned markers.
    \item \textbf{f02}: lymphoid contrast factor; B-cell versus T/NK separation.
    \item \textbf{f00}: strongest stage signal; granulocytic/T-NK developmental axis.
    \item \textbf{f03}: monocyte/macrophage versus granulocytic structure.
\end{itemize}
The same four-factor core persists in the hard-sparse surrogate (16 factors, 60 read/write genes), where it explains \(68.9\%\) of sparse-specific ablation impact. Read/write gene overlaps between the rank-64 and sparse factorizations confirm that the same mechanistic substrate survives aggressive compression (Supplementary~S27--S28).

\begin{table}[t]
\centering
\caption{Gene-program summary for the four core factors of the rank-64 compact operator. Ablation drop is the total pooled classification loss when the factor is removed. Top read/write genes and enriched marker sets are from cached local marker enrichment.}
\label{tab:gene_programs}
\footnotesize
\resizebox{\linewidth}{!}{
\begin{tabular}{clclp{4.2cm}}
\toprule
Factor & Role & Drop & Top read/write genes & Enriched programs \\
\midrule
\textbf{f01} & Branch routing & 0.430 & R: CSF3R, SLC6A1, IL7R; W: CSF3R, SLC25A37, FCER1G & Mono/macro diff.; na\"ive CD8$^+$ T; granulopoiesis \\[2pt]
\textbf{f02} & Lymphoid contrast & 0.254 & R: IL7R, GSTPL, IGF2R; W: HLA-DRA, MARCH6, IGKV5 & Lymphoid B vs T/NK; CD4$^+$ T maturation \\[2pt]
\textbf{f00} & Stage ordering & 0.280 & R: EPB41, GMR1, VPEL3; W: ISHC, LGALS14, EPB41 & Classical monocyte; na\"ive CD8$^+$ T program \\[2pt]
\textbf{f03} & Mono/macro struct. & 0.241 & R: LGALS3, GT58, LYZ; W: VPEL5, ALPL, SLC28A3 & Macrophage; granulocyte; B-cell markers \\
\bottomrule
\end{tabular}
}
\end{table}

\begin{figure}[t]
    \centering
    \includegraphics[width=0.85\linewidth]{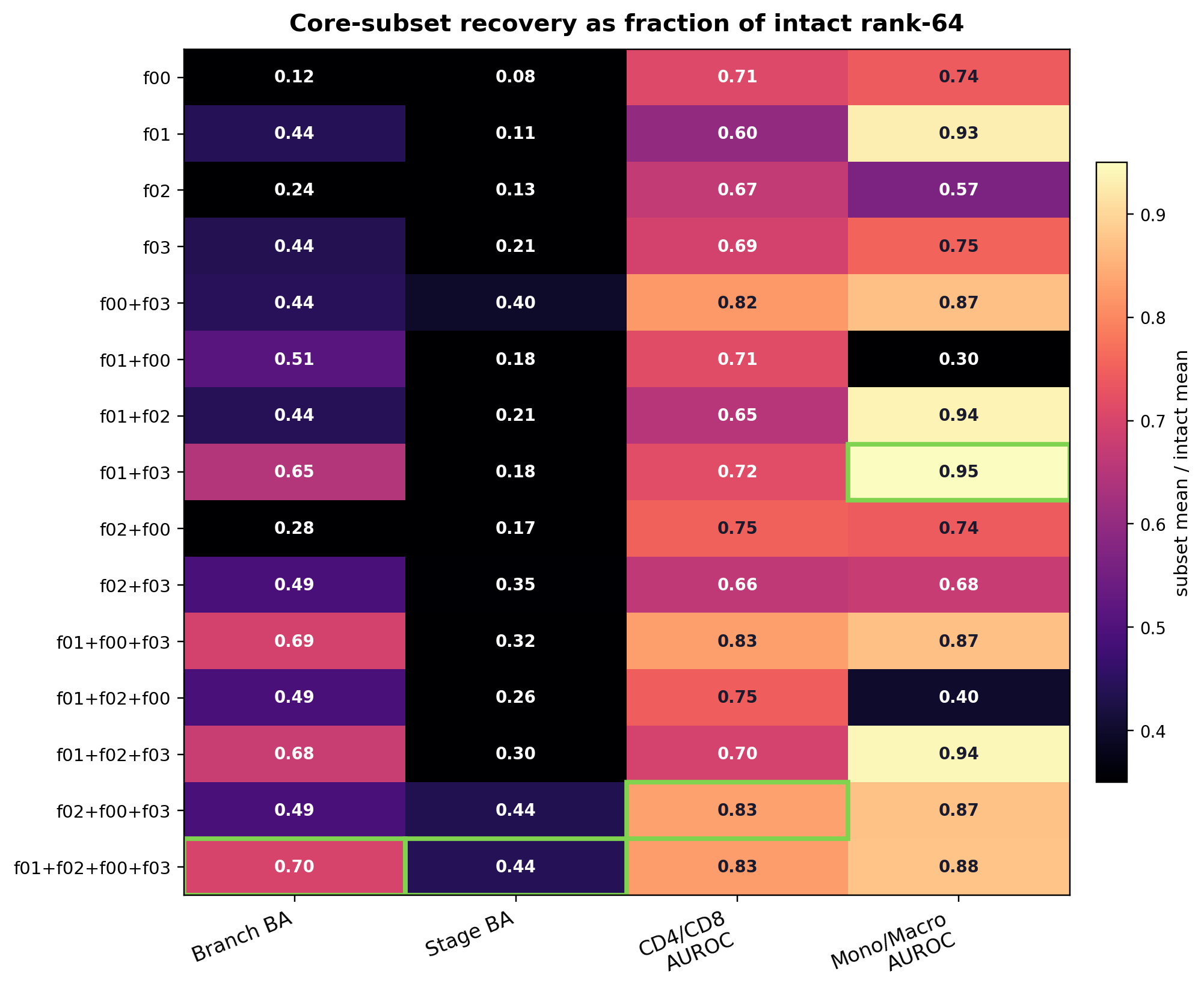}
    \caption{Core-subset recovery as a fraction of intact rank-64 performance across the exhaustive 15-subset interaction sweep. Green borders mark the best subset for each endpoint. No subset closes the gap to the intact model (all remain BH-significant); task-specialized circuitry is visible---mono/macro is well-recovered by the pair \{f01,\,f03\} alone (0.95$\times$), while branch and stage require all four factors. Best subsets per endpoint are listed in Supplementary Table~\ref{tab:rank64_core_interaction_best_subsets}.}
    \label{fig:core_interactions}
\end{figure}

Latent intervention tests on the manifold (protocol in Supplementary~S11) further support directional biological structure: stem\(\rightarrow\)myeloid intervention produced strong monotonic enrichment (\(\rho{=}0.928\), \(p{=}4.9{\times}10^{-6}\)), and stem\(\rightarrow\)erythroid showed a similarly strong shift (\(\rho{=}0.739\), \(p{=}0.0039\)). The CD4\(\rightarrow\)CD8 intervention, by contrast, was weak and non-significant (\(\rho{=}{-}0.518\), \(p{=}0.070\)), consistent with subtype-local complexity that resists simple latent-axis manipulation. Bootstrap and seed-perturbation audits confirmed stability of core claims (trustworthiness \(0.979\pm0.005\); CD4/CD8 AUC \(0.758\pm0.013\) across five seeds).

The 10D latent space distributes biological information across dimensions in a structured manner. Default 3D views under-represent T-cell subtype separability: CD4/CD8 AUROC is \(0.646\) in 3D but \(0.822\) in the full 10D latent, with the strongest single axis at dimension~4 (AUROC \(0.845\)). Monocyte/macrophage separation peaks at dimension~8 (AUROC \(0.907\)), and global stage/branch structure concentrates at dimension~9 (stage \(\eta^2{=}0.952\), branch \(\eta^2{=}0.883\), where \(\eta^2\) is the ANOVA effect size measuring the proportion of variance explained by the grouping variable). A guardrailed T-cell separation lens---a visualization-only rotation that replaces one 3D axis with the most discriminative latent dimension for CD4/CD8 separation, controlled by a regularization parameter \(\lambda{=}8\) that limits distortion of the global layout---improves CD4/CD8 AUROC from \(0.835\) to \(0.901\) while maintaining trustworthiness (methods in Supplementary~S9; results in Supplementary~S24). Flatness/ripple analysis (methods in Supplementary~S10) shows the manifold is predominantly sheet-like (best 2D plane explains \(99.25\%\) of 3D variance) with statistically significant residual structure (\(R^2{=}0.171\), \(p{=}0.0033\)), though residual-axis associations with curated erythroid programs did not survive BH-FDR correction (Supplementary~S25--S26).

\section{Discussion}

\paragraph{The manifold is real and transferable.}
The hematopoietic manifold discovered in scGPT is not an internal-only artifact. Strict non-overlap external transfer preserved geometric fidelity and holdout robustness, while a separate frozen-head zero-shot test on an independent multi-donor panel remained strongly significant. The lung panel failure under the same gates demonstrates that high global correlation alone is insufficient---robust transfer requires trustworthiness and clade holdout controls.

\paragraph{The extracted algorithm reshapes task geometry.}
The standalone benchmark clarifies what the extracted algorithm does well and where it does not lead. Pseudotime-depth alignment is the cleanest advantage: the cell-trained head significantly outperforms every tested alternative in paired split-level statistics. Classification leadership is mixed---the cell head leads on branch balanced accuracy and subtype AUROCs, but diffusion-style baselines and raw expression lead on some stage/branch macro-F1 endpoints. Importantly, the extraction gain is not reducible to a trivially better probe on frozen scGPT embeddings, nor to a small MLP trained directly on raw expression (Supplementary~S19--S22). The lightweight head reshapes the task geometry in ways not fully reproduced by either alternative, while requiring $34.5\times$ less compute and ${\sim}1{,}000\times$ fewer trainable parameters than the frozen-embedding MLP path.

\paragraph{Compaction localizes the signal.}
A single attention unit (L2H5) carries substantial transferable developmental geometry, and the compact operator based on this head preserves classification utility while shrinking the artifact from 17.5 to 5.9 MB. Further compression to rank-64 (0.73 MB) remains viable at moderate cost. This progression from distributed pooled heads to a single localized operator materially strengthens the extraction claim and connects to the broader paradigm of knowledge distillation \cite{hinton2015distilling} and the lottery ticket hypothesis \cite{frankle2019lottery}: a sparse subnetwork---here a single attention head---carries the essential computation, while the remaining components contribute incremental refinement rather than qualitatively new information.

\paragraph{Mechanistic interpretability reaches gene-program resolution.}
Factor ablation and sparse factorization of the compact operator resolve a four-factor core into explicit hematopoietic gene programs. This connects to the broader mechanistic interpretability program in language models, where circuit-level analysis \cite{olah2020zoom,elhage2021mathematical,conmy2023automated} and sparse dictionary learning \cite{bricken2023monosemanticity,templeton2024scaling} have revealed interpretable computational structure, including complete circuits for specific algorithmic tasks \cite{wang2023interpretability,neel2023progress} and methods for tracing causal pathways through sparse feature graphs \cite{marks2024sparse}. Our result represents a qualitative step beyond prior mechanistic interpretability of biological foundation models, which has typically stopped at attention pattern or embedding analyses \cite{kendiukhov2026attention}. While sparse autoencoder decomposition of these same models reveals rich biological feature organization but minimal causal regulatory logic \cite{kendiukhov2026sparse}, and causal circuit tracing shows convergent inhibitory-dominant architectures across model families \cite{kendiukhov2026causal}, the present work demonstrates that the structured knowledge these models \emph{do} encode---developmental geometry---can be extracted, compressed, and deployed as a competitive standalone algorithm. The core factors show task-specialized circuitry---mono/macro separation concentrates in just two factors (f01+f03), CD4/CD8 in triples centered on f00+f03, while branch and stage require the full four-factor core---rather than redundant additive structure.

\paragraph{A general method for extracting algorithms from biological foundation models.}
Beyond the specific hematopoietic result, this work demonstrates a reusable methodology for discovering and extracting biologically meaningful algorithms from the internals of biological foundation models. The three-stage extraction pipeline (operator export, learned adaptor, task readout) is model-agnostic: it requires only that the source model produces structured internal representations, and imposes no assumptions about the specific biological system. The extracted algorithms can be both more performant and dramatically faster than conventional alternatives---here, the extracted head significantly outperforms established trajectory-inference methods on pseudotime-depth alignment while completing evaluations \(34.5\times\) faster and requiring \({\sim}1{,}000\times\) fewer trainable parameters than the frozen-embedding probing path. This suggests that biological foundation models may harbor a library of compact, deployable algorithms that mechanistic interpretability can systematically surface, and that these algorithms need not be retrained or fine-tuned for each downstream application. The same protocol successfully recovered a second manifold family (intercellular communication, H38) under strict non-cheating external criteria: the final rescued variant achieved external correlation \(0.983\), internal trustworthiness \(0.908\), and a positive baseline margin (\(+0.006\) versus PCA-2; blocked-permutation \(p{=}0.0005\)), with axes resolving cytokine/chemokine versus lipid signaling programs (\(\eta^2{=}0.735\); Supplementary~S32--S37). Our work connects to a broader tradition of analysis methods for neural language processing \cite{belinkov2019analysis}, extending these approaches from linguistic to biological domains. Methodologically, the workflow moves mechanistic interpretability from qualitative inspection toward quantitative, falsifiable geometric auditing. More broadly, manifold geometry can serve as a hypothesis engine: branch placement, axis loadings, and residual ripple structure together nominate candidate regulatory transitions that can be prioritized for targeted perturbation or longitudinal validation.

\paragraph{Head-training granularity.}
The hybrid-topology sweep extends the extraction result into an explicit optimization frontier. A stronger topology prior yields markedly cleaner branch geometry but can degrade distance-alignment fidelity; a weaker prior preserves global alignment while still improving parts of geometric readability. The Pareto structure across head variants (anchor, cell, hybrid) suggests that no single objective simultaneously maximizes all downstream endpoints, and that the choice of training granularity should be guided by the intended application.

\paragraph{Low-rank and sparse compression boundaries.}
The low-rank follow-up sharpens the boundary. A third compression step via truncated-SVD surrogates does not preserve the near-lossless behavior of the dense single-head compact operator at very low rank. The best point, top1 + rank64, still beats frozen scGPT avg-pool + MLP on all eight pooled classification endpoints and shrinks the artifact further to \(0.73\) MB, but it incurs BH-significant losses versus dense compact top1 on 5/8 endpoints. More aggressive ranks degrade too much. The rank64 factor-ablation audit then shows why a further compression step remains plausible: necessity is highly concentrated, with the top four factors explaining \(66.2\%\) of total pooled ablation impact and many later factors being broad but weak stabilizers. The completed sparse follow-up closes that next question more sharply. A naive hard sparse surrogate can be made extremely small (\(1{,}920\) active operator weights, \(124{,}512\)-byte package) but falls back to baseline-like performance. The positive result is interpretability, not frontier accuracy.

\paragraph{Dimensionality and geometric fine structure.}
``Low-dimensional'' in this setting does not mean strictly 2D/3D. The data support a compact but higher-dimensional geometry: major lineage topology is visible in 3D, while full holdout-robust structure and subtype resolution are better captured in an \(8\)--\(10\)D latent manifold. The flatness/ripple analysis refines this view: 3D maps are sheet-like with structured residuals, and latent residual-axis tests show that fine geometric structure persists directly in LET-10D representation. Ripple semantics remain unresolved: we can detect structured residual geometry reliably, but mapping those residual axes to specific curated programs is not yet significant under multiple-testing control.

\paragraph{Limitations.}
External validation covers three panels (two passing, one failing); broader cross-cohort replication would strengthen generalization claims. Biological rulers (ontology distances, regulatory priors) are proxies---no ground-truth developmental metric exists. The benchmark panel, while strong, is not exhaustive, and classification leadership remains mixed across endpoints; claims should be interpreted as benchmark-specific. Donor reuse across the 88 Robust-V2 splits is substantial (15 unique test donors), limiting the independence of split-level inference. Low-rank compression below rank-64 degrades substantially, and sparse surrogates trade accuracy for interpretability rather than preserving the extracted-model advantage. Factor ablation uses a fixed-probe necessity protocol; it measures what the current model uses, not whether task performance could be recovered after retraining without a given factor.

\section{Conclusion}
We discovered a compact hematopoietic manifold in scGPT, validated it on external datasets in strict zero-shot mode, and extracted it as a standalone algorithm that significantly outperforms established methods on pseudotime-depth alignment while remaining competitive on classification. Multi-stage compaction reduced the operator to a single attention head (5.9 MB) without significant loss, and further to a rank-64 surrogate (0.73 MB) whose four-factor mechanistic core resolves into explicit hematopoietic gene programs. To our knowledge, this represents the first competitive, biologically useful algorithm extracted from a foundation model via mechanistic interpretability, and the most detailed mechanistic decomposition of such an extracted algorithm. The extraction method generalizes to a second biological manifold family (intercellular communication) and is applicable to other foundation models and developmental systems.

\clearpage
\section*{Supplementary Material}

\subsection*{S1. Autonomous research loop and hypothesis space}
The project used an autonomous multi-role loop centered on executor and reviewer/critic roles, with explicit hypothesis selection, gate auditing, and fast pivot rules. This main paper focuses on the validated hematopoiesis branch (H65) and its closure tests; a second validated branch (H38) is reported in the Supplementary Material after these sections, while other non-passing branches are included only as condensed context. The autonomous loop followed three stages under fixed gates. First, multiple exploratory branches were rejected because they failed effect-size or holdout robustness criteria. Second, H65 emerged as the first strong positive branch, while its paired null branch (H66) remained non-passing. Third, closure analyses (subspace/objective gap-fill, ablation+CI tests, equation-level LET replication, and strict external non-overlap transfer) elevated the H65 line to a closed, high-confidence result. Detailed iteration-by-iteration logs remain in \texttt{iterations/} in the project repo.

\subsection*{S2. Artifact-closure protocol}
Before interpreting any result, the loop enforced an artifact-closure pipeline:
\begin{itemize}[leftmargin=1.5em]
    \item \textbf{A: local path reconciliation} (canonical-path verification),
    \item \textbf{B: local materialization} (rebuild missing derived artifacts),
    \item \textbf{C: network retrieval} (autonomous download when local recovery fails),
    \item \textbf{D: schema enrichment} (metadata audit and harmonization).
\end{itemize}
For the final H65 cycle, A and D succeeded, B/C were not required, and no unresolved artifacts remained.

\subsection*{S3. Equation-level LET objective details}
The main text presents the core LET objective equations (Section~2.3). Here we document the shared quality gates applied across all panels:
\begin{itemize}[leftmargin=1.5em]
    \item trustworthiness \(\ge 0.80\),
    \item random holdout corr \(\ge 0.20\),
    \item donor holdout corr \(\ge 0.20\),
    \item clade/branch holdout corr \(\ge 0.20\).
\end{itemize}
These thresholds were fixed before any panel-specific evaluation and applied uniformly to internal, lung external, strict non-overlap external, and frozen zero-shot transfer panels. Positive claims were accepted only when all primary gates passed simultaneously.

\subsection*{S4. Frozen-head zero-shot transfer protocol}
For the transfer confirmation run, we used a strict \textbf{train-once, transfer-many} protocol:
\begin{enumerate}[leftmargin=1.5em]
    \item Train the LET head only on the internal H65 anchor panel and select the best internal dimension (\(d=10\)).
    \item Reconstruct that exact head deterministically (model seed \(15052\)); freeze all parameters.
    \item Build external anchors and sampled-cell views from a separate multi-donor immune panel (7 cohorts).
    \item Apply the frozen head directly (zero-shot; no external optimization/retraining).
    \item Evaluate alignment with the same blocked-permutation and trustworthiness controls.
\end{enumerate}
This protocol separates representation transfer from dataset-specific fitting, which is important for future applications.

\subsection*{S5. Cell-trained transferable head variant}
To test whether training granularity changes what the transferred manifold emphasizes, we trained an alternative internal head on individual cells (not anchors) and then transferred it zero-shot to the same strict non-overlap external anchors. The internal training set contained 6,925 mapped cells across 34 stages and 23 donors. We used a stage-balanced sampler (maximum 500 cells/stage) and optimized a weighted objective with four terms: stage-distance fitting, local-neighborhood preservation, reconstruction, and stage classification (\(w_{\text{stage}}=1.0\), \(w_{\text{local}}=0.1\), \(w_{\text{recon}}=0.08\), \(w_{\text{cls}}=0.4\); 120 epochs; batch size 896; \(d=10\)). No external rows were used during head training and no external fine-tuning was allowed at evaluation time.

\subsection*{S6. Hybrid topology-preserving variant}
We tested a hybrid head that augments the cell-level objective with an anchor-topology prior derived from the internal anchor LET-10D manifold. Let \(\mu_s(z)\) be the latent centroid for stage \(s\), and \(D_{\mathrm{ref}}\) the pairwise stage-centroid distance matrix from the internal anchor head (shared stages only). The hybrid objective is:
\[
\mathcal{L}_{\mathrm{hybrid}}=
\mathcal{L}_{\mathrm{cell}}
+ \lambda_{\mathrm{topo}}\lVert D(\mu(z)) - D_{\mathrm{ref}}\rVert^2
+ \lambda_{\mathrm{compact}}\sum_s \mathbb{E}_{i\in s}\lVert z_i-\mu_s(z)\rVert^2 .
\]
We ran two sweeps on the same strict non-overlap external evaluation panel:
\begin{itemize}[leftmargin=1.5em]
    \item \textbf{strong-topology sweep:} \(\lambda_{\mathrm{topo}}\in\{0.04,0.08,0.12,0.18\}\), \(\lambda_{\mathrm{compact}}=0.015\),
    \item \textbf{conservative sweep:} \(\lambda_{\mathrm{topo}}\in\{0.005,0.01,0.02,0.03\}\), \(\lambda_{\mathrm{compact}}=0\).
\end{itemize}
To choose a representative hybrid per sweep, we used a fixed external objective:
\[
\mathrm{score}= \rho_{\mathrm{resid}} + 0.25\,\mathrm{AUC}_{\mathrm{CD4/CD8}} + 0.25\,\mathrm{AUC}_{\mathrm{Mono/Macro}} + 0.40\,\mathrm{silhouette}_{\mathrm{branch}} - 0.20\,\mathrm{spread}_{\mathrm{norm}}.
\]

\subsection*{S7. Statistical controls and robustness tests}
We applied donor/tissue blocked permutations, confound-residualized correlations, branch holdouts, objective ablations, and bootstrap confidence intervals. Positive claims were accepted only when all primary gates passed. Bootstrap estimates were stable across core endpoints: CD4/CD8 AUC mean \(0.826\), 95\% CI \([0.772,0.876]\); branch silhouette mean \(0.404\), 95\% CI \([0.362,0.442]\); residual ripple \(R^2\) (PC4) mean \(0.274\), 95\% CI \([0.198,0.344]\). Across five seeds (\(\{42,43,44,45,46\}\)), trustworthiness remained high (\(0.979\pm0.005\)), while CD4/CD8 AUC varied moderately (\(0.758\pm0.013\)).

\subsection*{S8. Benchmark evaluation protocol details}
To quantify practical utility of extracted heads, we ran donor-level post-hoc benchmarks on the strict non-overlap external panel (616 anchors, 21 donors). The finalized Robust-V2 campaign used grouped donor-holdout splits with 100{,}000 training cells per split and strict test-only evaluation. It comprises three completed batches: one full 40-split run (all methods including scVI) and two additional 24-split core-only runs, for 88 pooled splits total.

We compared paper-referenced extracted variants (anchor head, cell head, hybrid-conservative head) against raw-expression, PCA/SVD, CellTypist, scVI-10D, Scanpy-DPT, and Palantir baselines. Metrics were branch and stage macro-F1 / balanced accuracy, CD4/CD8 and monocyte/macrophage AUROC / balanced accuracy, and pseudotime-depth Spearman correlation.

As a targeted ablation, we also compared learned heads against two native frozen scGPT embedding baselines: (i) whole-human average-pooled contextual token embeddings and (ii) continual-pretrained \texttt{<cls>} embeddings.

Classification endpoints were evaluated strictly on held-out test donors for all methods. For pseudotime, all methods were evaluated with a donor-local test-only protocol (pseudotime was computed separately within each test donor's cells), including donor-local native DPT/Palantir computation. We report both the primary signed pseudotime-depth Spearman and an orientation-aware summary (absolute Spearman and sign-share) to separate ordering strength from direction choice. Paired statistical comparisons were computed on split-level donor-mean differences using two-sided Wilcoxon signed-rank tests with BH-FDR correction.

\subsection*{S9. T-cell resolution lens methods}
A key interpretability issue was apparent CD4/CD8 mixing in default 3D projections. We tested whether this reflected absent signal or projection loss. Using strict non-overlap external anchors, we compared separability in displayed 3D versus full 10D LET latent space. We then defined a visualization-only lens: keep global \(x/y\), replace \(z\) by a discriminative latent-axis score (best CD4/CD8 dimension), preserving global context while exposing subtype separation.

\subsection*{S10. Flatness and ripple test methods}
To test whether our geometry resembles the Goodfire ``flat + ripples'' pattern, we ran two related analyses.

\textbf{Projection-level test (3D coordinates).} For each 3D manifold view, we fit a best 2D plane (weighted PCA), quantified residual structure (Moran's I on plane residuals), and scanned sinusoidal fits after quadratic detrending, with permutation tests.

\textbf{Latent-level test (LET-10D coordinates).} We reconstructed the exact LET-10D latent \(z\), defined a 2D substrate using latent PC1/PC2, then for each residual axis (PC3--PC10) tested detrended sinusoidal structure over in-plane directions with permutation tests and BH-FDR correction across residual axes.

We used a conservative residual-axis significance rule (\(q \le 0.05\), sinusoid \(R^2 \ge 0.10\), cycles-per-span \(\ge 1.5\)). We additionally ran a covariance-matched Gaussian null calibration to verify that the detector does not trigger on generic anisotropic clouds.

\subsection*{S11. Phase-2 deepening protocol}
To strengthen mechanistic interpretation beyond alignment and transfer, we ran a targeted second-pass protocol on the same H65 manifold:
\begin{itemize}[leftmargin=1.5em]
    \item \textbf{Latent interventions:} move source anchors along biologically motivated latent directions (stem\(\rightarrow\)erythroid, stem\(\rightarrow\)myeloid, CD4\(\rightarrow\)CD8), decode to feature space, and quantify monotonic target-fraction and graph-depth shifts.
    \item \textbf{Branchpoint quantification:} build stage-centroid kNN graphs, extract MST topology, and quantify root articulation, branchpoint count, endpoint reachability, and first-hop diversity from HSC.
    \item \textbf{Ripple semantics:} test whether significant residual ripple axes associate with curated biological program scores (Spearman + BH-FDR), then measure ripple \(R^2\) change after program residualization.
    \item \textbf{Stability envelope:} bootstrap confidence intervals for key outcomes (CD4/CD8 AUC, branch silhouette, residual ripple \(R^2\)) and seed-sensitivity audits across \(\{42,43,44,45,46\}\).
\end{itemize}

\subsection*{S12. Extraction pipeline terminology and compression stages}
Because later compression experiments build on the same exported object, we make the pipeline boundary explicit. The reported standalone method has three stages:
\begin{enumerate}[leftmargin=1.5em]
    \item \textbf{Direct operator extraction from scGPT.} Starting from the frozen whole-human checkpoint, we read native attention operators \(A_{\ell,h}\in\mathbb{R}^{1200\times 1200}\) and construct a fixed feature map. This stage is a direct checkpoint export: no target labels are used and no parameters are optimized.
    \item \textbf{Small learned adaptor on top of the extracted operator.} We train a lightweight head \(g_{\theta}\) on internal data only, mapping fixed extracted features to the task-agnostic manifold latent \(z\) (typically \(d=10\)). This is the only learned part of the shared representation.
    \item \textbf{Task-specific readout.} For downstream classification or regression, we train a separate probe \(h_{\phi}\) on top of \(z\). These probes are deliberately small and are not part of the extracted representation itself.
\end{enumerate}
Thus, the benchmarked representation should be described precisely as \emph{directly extracted scGPT operator + small learned adaptor}, not as a zero-learning export of the entire pipeline.

This distinction governs later compression stages. Compact-V1 remains a \emph{directly extracted operator} because it replaces the pooled drift map by a weighted sum of native top-ranked heads,
\[
\widetilde{A}_{k}=\sum_{i=1}^{k}\alpha_i A_{\ell_i,h_i},
\]
while still retraining only the small adaptor and downstream probe. By contrast, any subsequent low-rank, sparse, or quantized compression of \(\widetilde{A}_{k}\) should be interpreted as a \emph{compressed surrogate of the extracted operator}, not as a direct checkpoint export of new native scGPT parameters.

In the low-rank compression step, we fit truncated-SVD surrogates
\[
\widetilde{A}_{k}\approx U_rV_r^{\top}, \qquad r\in\{8,16,32,64\},
\]
for the compact top1/top2 operators and then retrain only the same lightweight 10D adaptor and downstream task probes. In the sparse follow-up, we applied a hard top-\(k\) pruning rule to the selected rank64 surrogate: retain only a subset of parent factors and keep only the top-\(k\) read genes and top-\(k\) write genes per retained factor. This produces an even smaller \emph{sparse surrogate of the low-rank surrogate}, again followed by retraining only the same adaptor and downstream probes.

For the factor-necessity audit, we took the selected rank64 surrogate and kept its downstream task probes fixed. Each leave-one-out ablation zeroed a single factor only at inference time while leaving all other factors and the trained probes unchanged. This fixed-probe protocol measures \emph{necessity of the existing factorization}, not recoverability after retraining; retraining after each factor drop would answer a different question.

\subsection*{S13. Trajectory and stage-order consistency}
Trajectory analyses support developmental ordering for several hematopoietic paths. In the expanded manifold, strongest significant signals were erythroid \(\rho=0.768, p=0.0017\), granulocytic \(\rho=0.568, p=0.0033\), and trunk HSC/HPC/CMP \(\rho=0.611, p=0.0183\). A coarse canonical myeloid trajectory remained positively associated (\(\rho=0.682, p=0.026\)), while fine monocyte\(\rightarrow\)macrophage ordering in the expanded panel was weak/non-significant (\(\rho=0.082, p=0.238\)). B\(\rightarrow\)plasma ordering was monotonic but small-sample limited (high \(\rho\), non-significant \(p\) in current anchor counts).

\begin{figure}[t]
    \centering
    \includegraphics[width=0.95\linewidth]{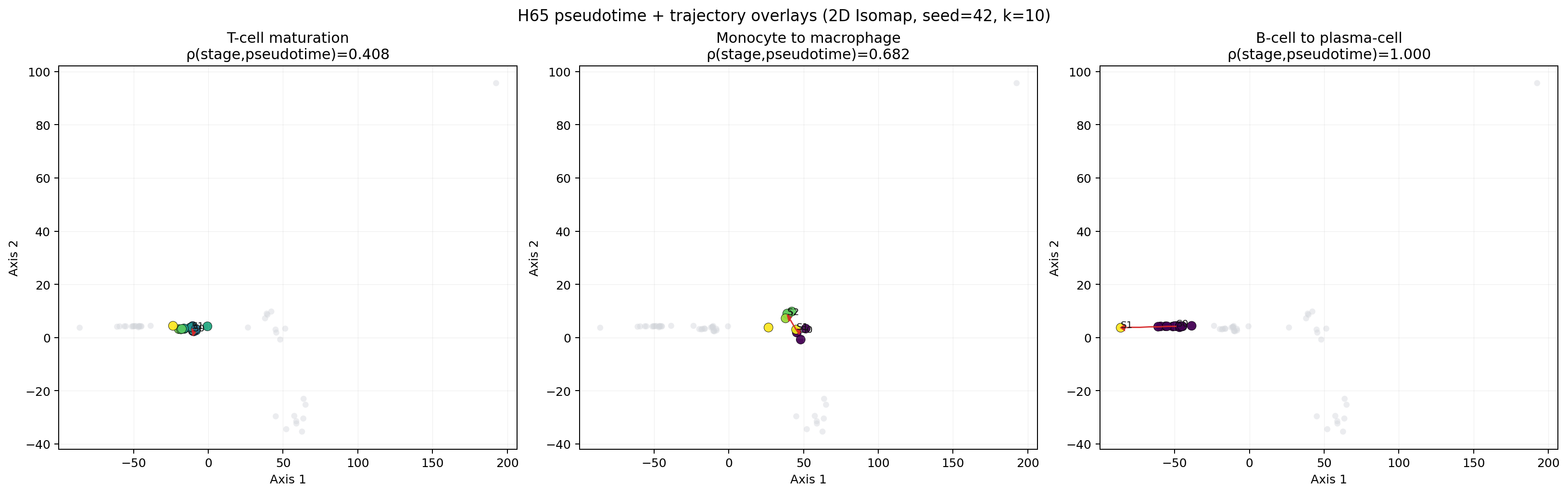}
    \caption{Pseudotime-annotated manifold view used for trajectory interpretation. The projection emphasizes developmental ordering from stem/progenitor regions toward branch-specific lineage endpoints.}
    \label{fig:trajectory_overlay}
\end{figure}

\subsection*{S14. Methodological closure and effective dimensionality}
\textbf{Flat-subspace training confirmation.} We verified that training the adaptor head in a flat (linear) subspace also passes transfer gates. The best flat-subspace dimension was \(d=8\), with random/donor/clade holdout correlation means of \(0.609/0.647/0.316\). The T/NK branch remained the weakest transfer target across all tested configurations.

\textbf{Objective and confidence-interval closure.} The best ablation configuration (combined distance and reconstruction objective at \(d=3\)) achieved random/donor/tissue/clade holdout means of \(0.637/0.610/0.571/0.301\), with bootstrap confidence intervals reported for each axis and improved T/NK transfer relative to the prior baseline.

\textbf{Effective dimensionality.} A three-dimensional projection is a visualization slice rather than the full manifold. In strict external non-overlap LET transfer, the selection objective increased consistently with dimensionality:
\begin{itemize}[leftmargin=1.5em]
    \item \(d=3\): objective \(0.507\), trustworthiness \(0.946\),
    \item \(d=6\): objective \(0.604\), trustworthiness \(0.976\),
    \item \(d=10\): objective \(0.632\), trustworthiness \(0.985\),
\end{itemize}
with all holdout gates passing at each tested \(d\). In explicit flat-subspace training, the best objective was at \(d=8\) (objective \(0.475\)). Together these results support an effective hematopoietic manifold size of approximately \(8\)--\(10\) dimensions.

\textbf{Internal LET confirmation.} The internal latent embedding transfer fully passed all quality gates (Table~\ref{tab:transfer}), confirming that the transfer protocol is consistent between internal and external panels.

\subsection*{S15. Anchor-trained versus cell-trained head comparison}
Using the same strict non-overlap external anchor panel (616 anchors), we compared the original anchor-trained transferable head with the cell-trained transferable head (Table~\ref{tab:anchor_vs_cell_head}). The cell-trained head slightly improved alignment metrics (corr \(0.607\rightarrow0.614\), residualized corr \(0.605\rightarrow0.612\)) and strongly improved global 3D subtype separability (CD4/CD8 AUC \(0.744\rightarrow0.886\), monocyte/macrophage AUC \(0.759\rightarrow0.937\)). Trustworthiness remained high but decreased slightly (\(0.972\rightarrow0.969\)).

The tradeoff appears in global display geometry: normalized within-stage spread increased (\(0.217\rightarrow0.314\)), branch silhouette decreased (\(0.013\rightarrow0.008\)), and a stem-centric radial proxy changed sign (stem-to-plasma minus stem-to-B distance \(+95.740\rightarrow-3.844\)). Cell-level training sharpens subtype margins but relaxes the compact branch-like arrangement seen in the anchor-trained map.

\begin{table}[t]
\centering
\caption{Anchor-trained vs cell-trained head on the same strict non-overlap external panel (global 3D view unless noted).}
\label{tab:anchor_vs_cell_head}
\small
\begin{tabular}{lcc}
\toprule
Metric & Anchor-trained & Cell-trained \\
\midrule
Corr (projected distance vs target) & 0.607 & 0.614 \\
Residualized corr & 0.605 & 0.612 \\
Trustworthiness & 0.972 & 0.969 \\
CD4/CD8 AUC(abs), global 3D & 0.744 & 0.886 \\
Monocyte/Macrophage AUC(abs), global 3D & 0.759 & 0.937 \\
Normalized within-stage spread & 0.217 & 0.314 \\
Branch silhouette mean (3D) & 0.013 & 0.008 \\
Stem\(\rightarrow\)(plasma minus B-cell) distance & \(+95.740\) & \(-3.844\) \\
\bottomrule
\end{tabular}
\end{table}

\subsection*{S16. Hybrid topology-preserving sweep}
The hybrid runs defined a clear Pareto-like frontier between developmental branch readability and stage-distance alignment. In the \textbf{strong-topology sweep}, the best run (\(\lambda_{\mathrm{topo}}=0.08\)) produced the clearest branch geometry (branch silhouette \(0.102\); stem\(\rightarrow\)(plasma minus B-cell) radial proxy \(+4.167\)), and retained strong subtype separation (CD4/CD8 AUC \(0.934\), monocyte/macrophage AUC \(0.914\)). The tradeoff was reduced global alignment (corr\(_{\mathrm{resid}}=0.472\)).

In the \textbf{conservative sweep}, the best run (\(\lambda_{\mathrm{topo}}=0.005\)) preserved alignment (corr \(0.634\), corr\(_{\mathrm{resid}}=0.632\)) while keeping strong CD4/CD8 separability (\(0.911\)); monocyte/macrophage separability dropped relative to pure cell-trained transfer (\(0.851\) vs \(0.937\)), and topology gains were more modest (branch silhouette \(0.037\)).

\begin{table}[t]
\centering
\caption{Transfer-head tradeoff frontier on strict non-overlap external anchors (global 3D metrics).}
\label{tab:hybrid_tradeoff}
\small
\resizebox{\linewidth}{!}{
\begin{tabular}{lcccc}
\toprule
Metric & Anchor-trained & Cell-trained & Hybrid strong-topology & Hybrid conservative \\
\midrule
Corr & 0.607 & 0.614 & 0.473 & 0.634 \\
Residualized corr & 0.605 & 0.612 & 0.472 & 0.632 \\
Trustworthiness & 0.972 & 0.969 & 0.969 & 0.959 \\
CD4/CD8 AUC(abs), global 3D & 0.744 & 0.886 & 0.934 & 0.911 \\
Monocyte/Macrophage AUC(abs), global 3D & 0.759 & 0.937 & 0.914 & 0.851 \\
Normalized within-stage spread & 0.217 & 0.314 & 0.348 & 0.318 \\
Branch silhouette mean (3D) & 0.013 & 0.008 & 0.102 & 0.037 \\
Stem\(\rightarrow\)(plasma minus B-cell) distance & \(+95.740\) & \(-3.844\) & \(+4.167\) & \(-0.397\) \\
\bottomrule
\end{tabular}
}
\end{table}

\subsection*{S17. Metric-level paired significance, orientation summary, and classification deltas}
\label{sec:s17}

\begin{table}[t]
\centering
\caption{Robust-V2 pseudotime orientation summary across pooled donor rows. Signed \(\rho\) is the primary metric; \(|\rho|\) and positive-sign share are orientation-aware summaries.}
\label{tab:posthoc_pt_orientation_v2}
\small
\resizebox{\linewidth}{!}{
\begin{tabular}{lcccc}
\toprule
Method & \(n_{\mathrm{rows}}\) & Mean signed \(\rho\) & Mean \(|\rho|\) & Positive-share (\%) \\
\midrule
H65 anchor head (10D) & 307 & 0.093 & 0.235 & 65.1 \\
H65 cell head (10D) & 352 & 0.249 & 0.439 & 71.0 \\
H65 hybrid conservative (10D) & 307 & 0.132 & 0.364 & 64.5 \\
Raw baseline (1200 log1p) & 307 & -0.044 & 0.266 & 37.5 \\
PCA-10 baseline & 307 & -0.007 & 0.259 & 38.1 \\
SVD-10 baseline & 307 & 0.002 & 0.279 & 51.8 \\
CellTypist stage-probability baseline & 306 & -0.064 & 0.253 & 41.5 \\
scVI-10D baseline & 125 & -0.103 & 0.274 & 37.6 \\
Scanpy-DPT baseline & 307 & -0.089 & 0.261 & 34.9 \\
Palantir baseline & 275 & -0.049 & 0.331 & 47.3 \\
\bottomrule
\end{tabular}
}
\end{table}

The orientation-aware view preserves the same ranking at the top: the cell head has both the highest signed mean and the highest absolute mean (\(|\rho|=0.439\)). Several baselines show non-trivial absolute association (for example Palantir \(|\rho|=0.331\)) but with weaker sign stability (positive-share near 35--52\%), consistent with frequent orientation reversals across donor-local evaluations.

\begin{table}[t]
\centering
\caption{Robust-V2 metric-level paired significance summary versus H65 cell head (comparator minus cell head; higher is better for all metrics).}
\label{tab:robustv2_allmetrics_pairwise_summary}
\small
\resizebox{\linewidth}{!}{
\begin{tabular}{lcccc}
\toprule
Metric & Cell head mean & Best comparator (mean) & Mean diff & Wilcoxon BH-\(q\) \\
\midrule
Branch balanced accuracy & 0.828 & Raw baseline (0.826) & -0.0018 & \(1.5\times10^{-1}\) \\
Branch macro-F1 & 0.584 & Palantir baseline (0.621) & +0.0358 & \(1.4\times10^{-10}\) \\
Stage balanced accuracy & 0.552 & Raw baseline (0.567) & +0.0148 & \(1.8\times10^{-7}\) \\
Stage macro-F1 & 0.216 & Palantir baseline (0.242) & +0.0254 & \(1.3\times10^{-6}\) \\
CD4/CD8 AUROC & 0.867 & H65 hybrid conservative (0.858) & -0.0088 & \(1.1\times10^{-15}\) \\
CD4/CD8 balanced accuracy & 0.798 & H65 hybrid conservative (0.794) & -0.0025 & \(1.3\times10^{-2}\) \\
Mono/Macro AUROC & 0.951 & H65 hybrid conservative (0.949) & -0.0014 & \(7.1\times10^{-3}\) \\
Mono/Macro balanced accuracy & 0.864 & PCA-10 baseline (0.871) & +0.0075 & \(1.0\times10^{-8}\) \\
Pseudotime-depth Spearman & 0.249 & H65 hybrid conservative (0.132) & -0.0561 & \(2.7\times10^{-7}\) \\
\bottomrule
\end{tabular}
}
\end{table}

This pooled paired view confirms a mixed classification profile and a strong pseudotime profile. The cell head is significantly better than the strongest comparator on CD4/CD8 AUROC, CD4/CD8 balanced accuracy, mono/macro AUROC, and pseudotime-depth Spearman; branch balanced accuracy is numerically close and not significantly different at BH-\(q\le0.05\). In contrast, branch macro-F1, stage balanced accuracy, stage macro-F1, and mono/macro balanced accuracy are significantly led by specific baselines.

\begin{figure}[p]
    \centering
    \includegraphics[width=\linewidth]{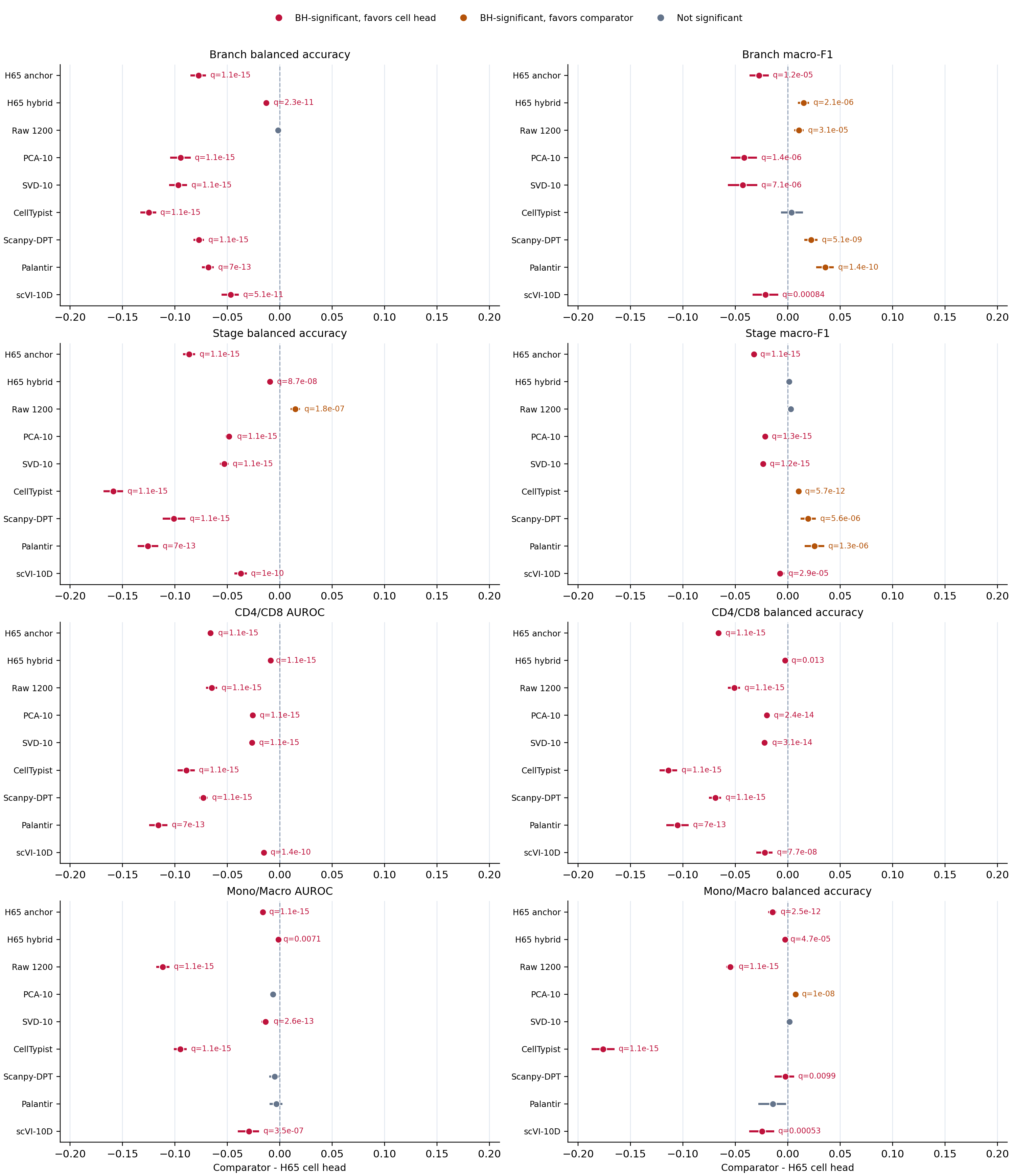}
    \caption{Robust-V2 splitwise paired deltas versus the H65 cell head for the eight non-pseudotime endpoints. Each panel shows comparator-minus-cell-head mean differences with 95\% bootstrap confidence intervals across grouped donor-holdout splits. Negative values favor the cell head. Red points indicate BH-significant cell-head advantages; gold points indicate BH-significant comparator advantages.}
    \label{fig:robustv2_classification_deltas}
\end{figure}

\textbf{Sign interpretation note.} Pseudotime-depth Spearman in Table~\ref{tab:posthoc_largecell_iter1} is the raw signed correlation \(\rho=\mathrm{Spearman}(\text{inferred pseudotime},\text{ontology depth})\), not a baseline-normalized score. Positive values indicate alignment with increasing developmental depth; negative values indicate inverse ordering. Negative signed \(\rho\) does not necessarily imply absent trajectory signal, because pseudotime orientation can be reversed by root choice or manifold direction; Table~\ref{tab:posthoc_pt_orientation_v2} therefore reports orientation-aware summaries.

\paragraph{Scope closure and residual caveats in Robust V2.}
The prior scope mismatch for native diffusion pseudotime is closed: all methods were evaluated donor-locally on test-only cells. Residual limitations remain. Test donor count per split is fixed at 4, and donor reuse is substantial (15 unique test donors across 88 pooled splits; 14 unique donors in the strict 40-split full-method subset). Method coverage is uneven for some endpoints (pseudotime rows: Palantir 70 splits, scVI 36 splits). To stress fairness, we additionally inspected the strict 40-split full-method subset, where the same pseudotime ordering persisted (cell head \(0.236\), scVI \(-0.103\), all paired deltas negative).

\subsection*{S18. Unsupervised branch-inference benchmark}
Because branch classification only tests label decodability, we added a separate branch-inference readout on the canonical 100k donor-holdout split (100k train, 139,357 test cells, 4 held-out donors). Each representation is passed to the same unsupervised branch wrapper without training a supervised branch classifier, and the resulting partitions are compared against annotation-derived reference branch groups.

\begin{table}[t]
\centering
\caption{Canonical 100k split unsupervised branch-inference benchmark using Scanpy \cite{wolf2018scanpy} neighbors\(+\)Leiden \cite{traag2019louvain}\(+\)PAGA. Values are donor-level means across all 4 held-out donors.}
\label{tab:branch_scanpy_canonical100k}
\small
\resizebox{\linewidth}{!}{
\begin{tabular}{lcccc}
\toprule
Method & ARI & AMI & NMI & V-measure \\
\midrule
H65 anchor head (10D) & 0.269 & 0.546 & 0.547 & 0.547 \\
H65 cell head (10D) & 0.330 & 0.520 & 0.521 & 0.521 \\
H65 hybrid conservative (10D) & 0.387 & 0.618 & 0.619 & 0.619 \\
Raw baseline (1200 log1p) & 0.393 & 0.626 & 0.627 & 0.627 \\
PCA-10 baseline & 0.305 & 0.581 & 0.582 & 0.582 \\
SVD-10 baseline & 0.326 & 0.591 & 0.592 & 0.592 \\
Frozen scGPT avg-pool & 0.302 & 0.555 & 0.557 & 0.557 \\
Frozen scGPT CP-\texttt{<cls>} & 0.315 & 0.574 & 0.576 & 0.576 \\
\bottomrule
\end{tabular}
}
\end{table}

On the full-coverage Scanpy-PAGA-Leiden benchmark, the raw-expression baseline is best on all four partition-agreement metrics, with the hybrid head close behind; the cell head is intermediate rather than dominant. This differs from the pseudotime benchmark, where the cell head is clearly best. The interpretation is that branch identity is highly decodable from the extracted head, but unsupervised coarse branch partition recovery is still strongest in raw or hybrid geometries under the current wrapper.

\begin{table}[t]
\centering
\caption{Canonical 100k split unsupervised branch-inference benchmark using Palantir fate assignments after a fair-scope rerun with frozen donor-local evaluation subsets and explicit audit logging.}
\label{tab:branch_palantir_canonical100k}
\small
\resizebox{\linewidth}{!}{
\begin{tabular}{lccccc}
\toprule
Method & \(n_d\) & ARI & AMI & NMI & V-measure \\
\midrule
H65 anchor head (10D) & 2 & 0.613 & 0.670 & 0.670 & 0.670 \\
H65 cell head (10D) & 4 & 0.120 & 0.293 & 0.294 & 0.294 \\
H65 hybrid conservative (10D) & 3 & 0.255 & 0.410 & 0.410 & 0.410 \\
Raw baseline (1200 log1p) & 3 & 0.338 & 0.437 & 0.437 & 0.437 \\
PCA-10 baseline & 2 & 0.145 & 0.288 & 0.288 & 0.288 \\
SVD-10 baseline & 3 & 0.358 & 0.461 & 0.462 & 0.462 \\
Frozen scGPT avg-pool & 1 & 0.198 & 0.353 & 0.353 & 0.353 \\
Frozen scGPT CP-\texttt{<cls>} & 2 & 0.300 & 0.334 & 0.335 & 0.335 \\
\bottomrule
\end{tabular}
}
\end{table}

\begin{figure}[t]
    \centering
    \includegraphics[width=\linewidth]{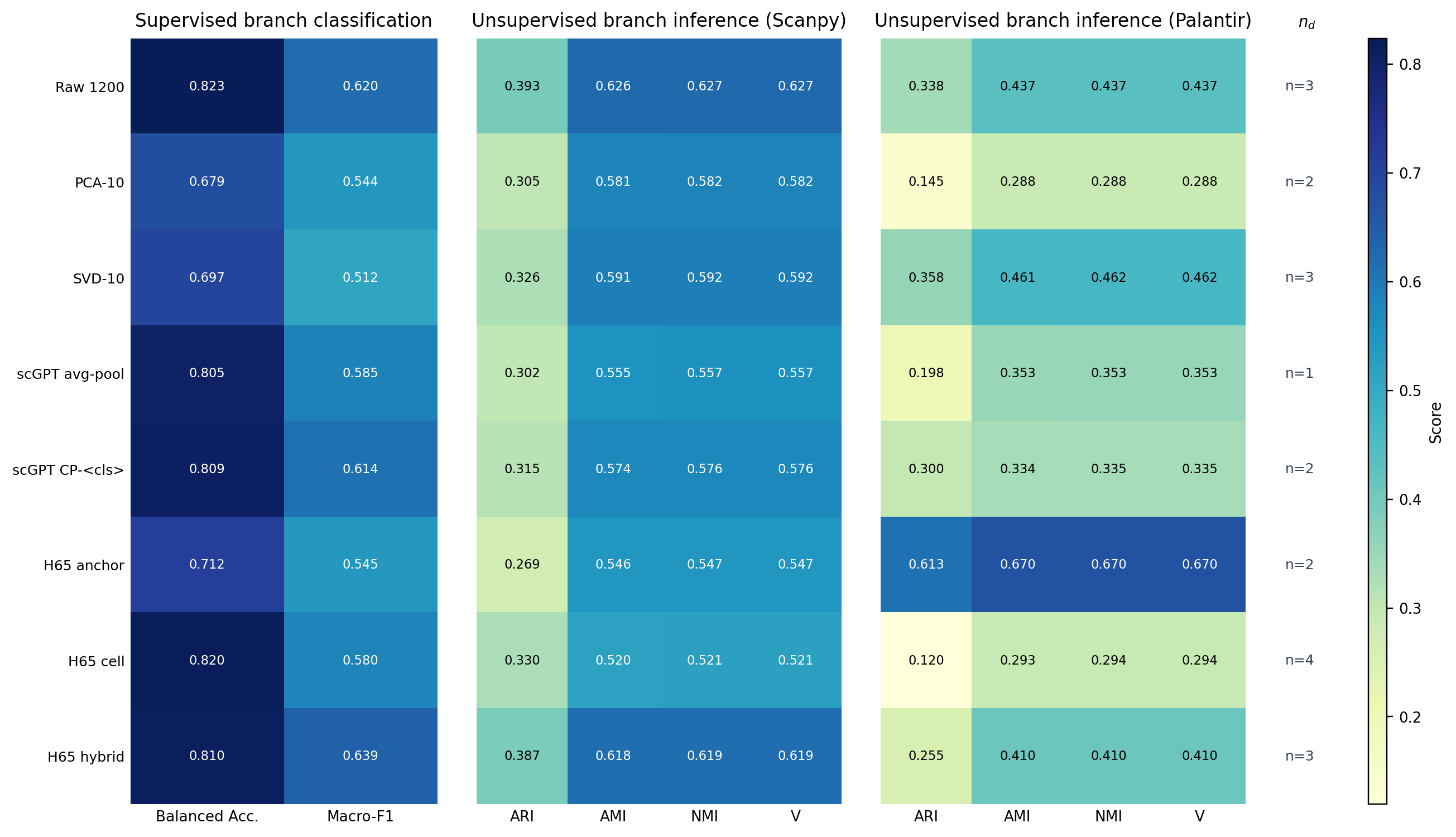}
    \caption{Supervised branch classification versus unsupervised branch recovery on the canonical 100k donor-holdout split. The task distinction is visible: branch information can be decodable under supervision even when unsupervised branch recovery remains wrapper- and coverage-sensitive.}
    \label{fig:branch_supervised_vs_unsupervised}
\end{figure}

\begin{figure}[t]
    \centering
    \includegraphics[width=\linewidth]{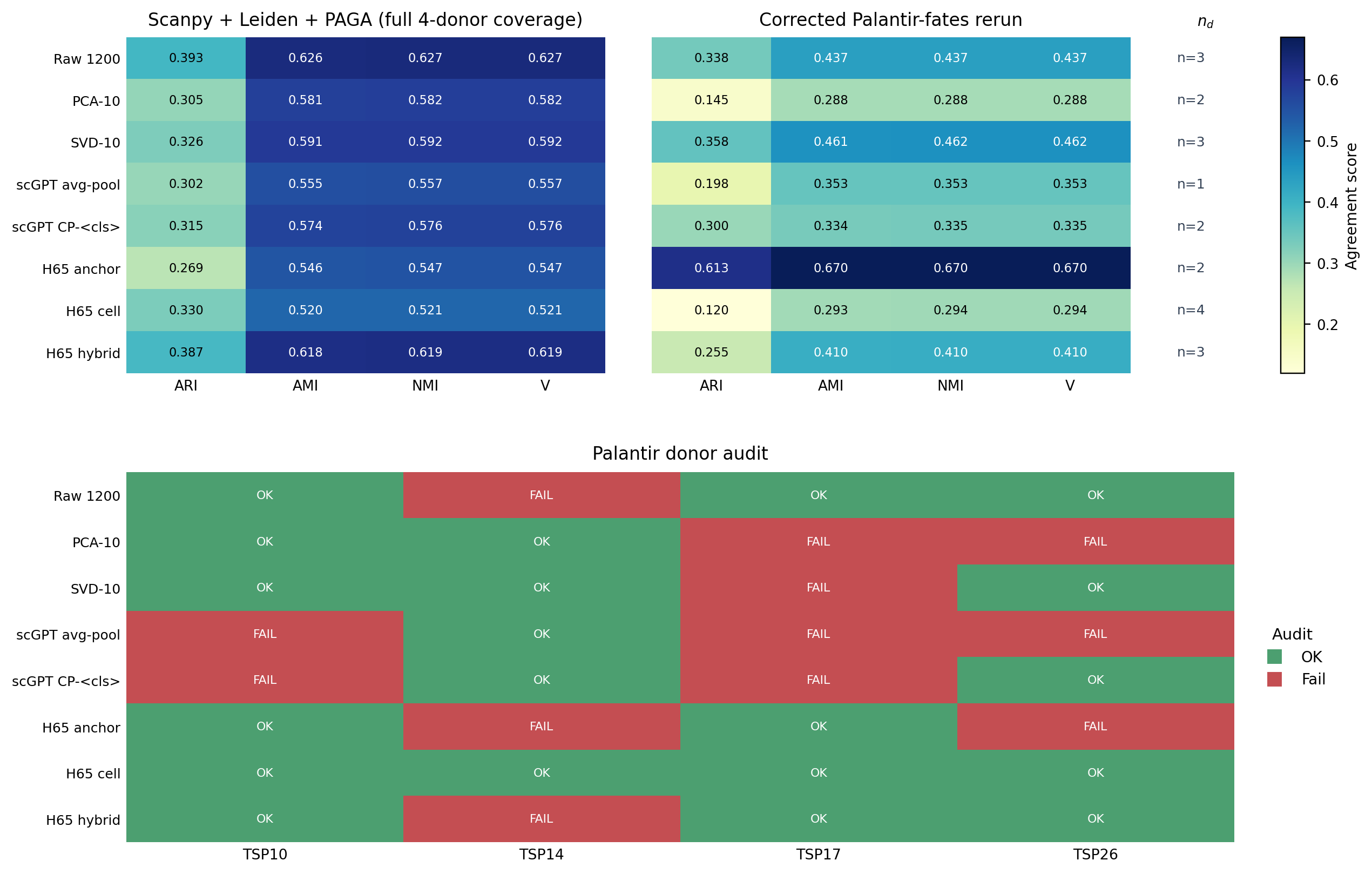}
    \caption{Branch-inference wrapper comparison. The Scanpy wrapper yields stable full-coverage rankings, whereas Palantir's strongest mean scores are tied to reduced donor support.}
    \label{fig:branch_wrapper_comparison}
\end{figure}

\textbf{Palantir audit.} The corrected Palantir branch-inference rerun evaluated the same held-out donors for all methods: \{TSP10, TSP14, TSP17, TSP26\}. However, valid Palantir outputs remained method-dependent because some donor/method pairs failed to produce at least two terminal states or encountered numerical errors. Table~\ref{tab:palantir_audit} lists the donor-level success matrix.

\begin{table}[ht]
\centering
\caption{Palantir donor-level success matrix on the canonical 100k donor-holdout split.}
\label{tab:palantir_audit}
\small
\begin{tabular}{lcccc}
\toprule
Method & TSP10 & TSP14 & TSP17 & TSP26 \\
\midrule
Raw baseline (1200 log1p) & ok & fail & ok & ok \\
PCA-10 baseline & ok & ok & fail & fail \\
SVD-10 baseline & ok & ok & fail & ok \\
Frozen scGPT avg-pool & fail & ok & fail & fail \\
Frozen scGPT CP-\texttt{<cls>} & fail & ok & fail & ok \\
H65 anchor head (10D) & ok & fail & ok & fail \\
H65 cell head (10D) & ok & ok & ok & ok \\
H65 hybrid conservative (10D) & ok & fail & ok & ok \\
\bottomrule
\end{tabular}
\end{table}

Across the 32 Palantir donor/method jobs, 20 succeeded and 12 failed. Failures were dominated by ``Palantir produced \(< 2\) terminal states'' (10 cases), with one infinity-bearing run and one eigensolver non-convergence. Because no donor succeeded for all methods simultaneously, the strict all-method shared-donor intersection was empty. We therefore treat Palantir as a qualitative stress test rather than a decisive ranking benchmark.

\subsection*{S19. Embedding ablation}
To determine whether the extracted head adds value beyond what is already present in scGPT's native embeddings, we compared it against two frozen scGPT embedding readouts: the average-pooled contextual embedding (averaging over all token positions) and the \texttt{<cls>}-token embedding (a single summary vector from the model's cell-processing checkpoint).

\begin{table}[t]
\centering
\caption{Embedding-ablation comparison on the held-out donor split (20k train, 30,137 test). Values are donor-level means.}
\label{tab:embedcmp_scgpt_vs_head}
\small
\begin{tabular}{lccc}
\toprule
Metric & H65 cell head (10D) & scGPT avg-pool & scGPT CP-\texttt{<cls>} \\
\midrule
Branch balanced accuracy (\(n_d=4\)) & 0.866 & 0.880 & 0.861 \\
Branch macro-F1 (\(n_d=4\)) & 0.529 & 0.526 & 0.503 \\
Stage balanced accuracy (\(n_d=4\)) & 0.610 & 0.593 & 0.570 \\
Stage macro-F1 (\(n_d=4\)) & 0.188 & 0.154 & 0.145 \\
CD4/CD8 AUROC (\(n_d=3\)) & 0.890 & 0.825 & 0.828 \\
Mono/Macro AUROC (\(n_d=2\)) & 0.926 & 0.798 & 0.775 \\
Pseudotime-depth Spearman (\(n_d=4\)) & 0.011 & -0.071 & -0.044 \\
\bottomrule
\end{tabular}
\end{table}

The cell-trained head outperforms both frozen embedding baselines on 6/7 metrics versus avg-pool and on all 7/7 versus CP-\texttt{<cls>}. The largest gains are on subtype-sensitive endpoints (CD4/CD8 and mono/macro AUROC) and stage-focused classification, while branch-level separability is already comparatively strong in native embeddings. Paired donor-level tests are directionally consistent but none survive BH-FDR (\(q>0.05\)) because per-metric donor counts are small (\(n_d=2\) to 4). We therefore treat this as targeted supporting evidence.

\subsection*{S20. Stronger frozen-encoder baseline (4D and 10D comparisons)}
\textbf{4D extracted head versus frozen scGPT + MLP.} To test a stronger direct-readout baseline under repeated-split statistics, we ran a grouped donor-holdout campaign with 12 splits comparing a 4D H65 cell head against frozen whole-human scGPT average-pooled embeddings with task-specific 3-layer MLP probes (\(512\rightarrow256\rightarrow128\rightarrow64\rightarrow C\), dropout \(0.1\)).

\begin{table}[t]
\centering
\caption{Repeated grouped-split comparison between a 4D extracted H65 cell head with linear probes and frozen scGPT avg-pool embeddings with 3-layer MLP probes (\(n_{\mathrm{split}}=12\)). Negative deltas favor the extracted head.}
\label{tab:head4d_vs_frozen_scgpt_mlp_metrics}
\small
\resizebox{\linewidth}{!}{
\begin{tabular}{lcccccc}
\toprule
Metric & \(n_{\mathrm{obs}}\) & H65 cell head (4D) + linear & Frozen scGPT avg-pool + MLP & \(\Delta_{\mathrm{base-head}}\) & 95\% CI & BH-\(q\) \\
\midrule
Branch macro-F1 & 40 & \textbf{0.597} & 0.547 & -0.045 & [-0.067, -0.023] & 0.010 \\
Branch balanced accuracy & 40 & 0.801 & \textbf{0.810} & 0.010 & [-0.001, 0.024] & 0.192 \\
Stage macro-F1 & 40 & \textbf{0.236} & 0.157 & -0.078 & [-0.087, -0.068] & 0.0013 \\
Stage balanced accuracy & 40 & \textbf{0.540} & 0.403 & -0.127 & [-0.164, -0.086] & 0.0020 \\
CD4/CD8 AUROC & 29 & \textbf{0.845} & 0.817 & -0.026 & [-0.033, -0.018] & 0.0013 \\
CD4/CD8 balanced accuracy & 29 & \textbf{0.780} & 0.744 & -0.034 & [-0.045, -0.025] & 0.0013 \\
Mono/Macro AUROC & 28 & \textbf{0.953} & 0.946 & -0.005 & [-0.014, 0.004] & 0.279 \\
Mono/Macro balanced accuracy & 28 & \textbf{0.880} & 0.844 & -0.034 & [-0.069, -0.001] & 0.119 \\
\bottomrule
\end{tabular}
}
\end{table}

The 4D head improves pooled metrics on 7/8 endpoints, with 5/8 advantages remaining BH-significant after paired splitwise correction. This rules out the alternative explanation that the head only looked favorable when native embeddings were paired with simplistic linear probes.

\begin{figure}[t]
    \centering
    \includegraphics[width=\linewidth]{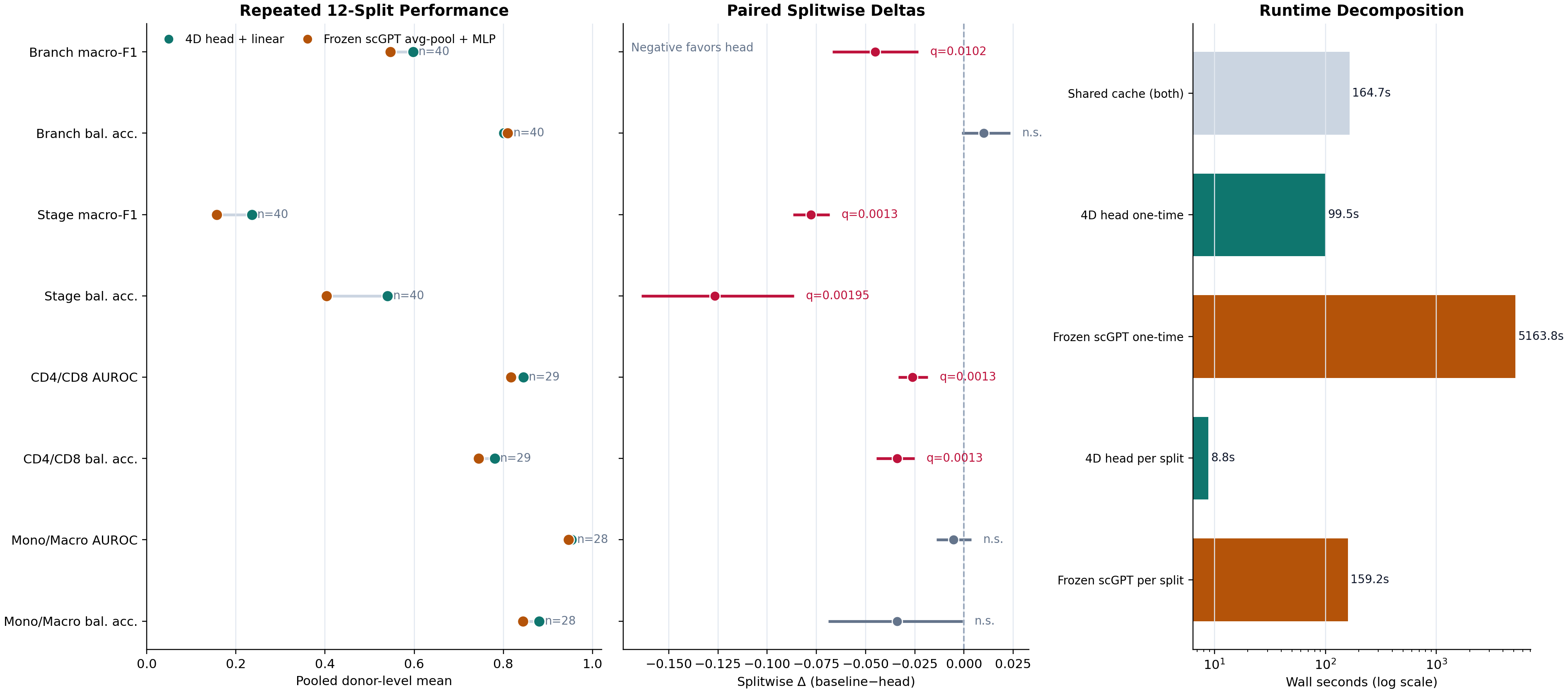}
    \caption{Repeated 12-split comparator between a 4D extracted head and frozen scGPT avg-pool embeddings with 3-layer MLP probes. Left: pooled means. Middle: paired splitwise deltas. Right: runtime decomposition.}
    \label{fig:head4d_vs_frozen_scgpt_robust}
\end{figure}

\textbf{10D extracted head + 2-layer MLP versus frozen scGPT + MLP.} We then tightened the comparison using a 10D H65 cell head with a 2-layer task probe (\(10\rightarrow64\rightarrow C\)).

\begin{table}[t]
\centering
\caption{Repeated grouped-split comparison between a 10D extracted H65 cell head with 2-layer MLP probes and frozen scGPT avg-pool + 3-layer MLP probes (\(n_{\mathrm{split}}=12\)). Positive deltas favor the extracted head.}
\label{tab:head10d_mlp2_vs_frozen_scgpt_mlp_metrics}
\small
\resizebox{\linewidth}{!}{
\begin{tabular}{lcccccc}
\toprule
Metric & \(n_{\mathrm{obs}}\) & H65 cell head (10D) + MLP2 & Frozen scGPT avg-pool + MLP & \(\Delta_{\mathrm{head-base}}\) & 95\% CI & BH-\(q\) \\
\midrule
Branch balanced accuracy & 40 & \textbf{0.827} & 0.810 & 0.017 & [0.012, 0.024] & 0.0011 \\
Branch macro-F1 & 40 & \textbf{0.556} & 0.547 & 0.010 & [-0.003, 0.022] & 0.194 \\
Stage balanced accuracy & 40 & \textbf{0.468} & 0.403 & 0.063 & [0.048, 0.077] & 0.0011 \\
Stage macro-F1 & 40 & \textbf{0.180} & 0.157 & 0.024 & [0.020, 0.027] & 0.0011 \\
CD4/CD8 AUROC & 29 & \textbf{0.856} & 0.817 & 0.038 & [0.032, 0.044] & 0.0011 \\
CD4/CD8 balanced accuracy & 29 & \textbf{0.779} & 0.744 & 0.034 & [0.028, 0.040] & 0.0011 \\
Mono/Macro AUROC & 28 & \textbf{0.956} & 0.946 & 0.010 & [0.004, 0.017] & 0.031 \\
Mono/Macro balanced accuracy & 28 & \textbf{0.865} & 0.844 & 0.021 & [0.002, 0.042] & 0.124 \\
\bottomrule
\end{tabular}
}
\end{table}

The 10D+MLP2 variant exceeds the frozen-scGPT baseline on 7/8 pooled metrics with 6/8 BH-significant after paired splitwise correction.

\subsection*{S21. Probe-depth comparison}
We closed the downstream probe-depth question on the broader 40-split Robust-V2 family. We evaluated the robust linear extracted head against both a modest non-linear readout (MLP2, \(10\rightarrow64\rightarrow C\)) and a deeper alternative (MLP3, \(10\rightarrow128\rightarrow64\rightarrow C\)). Neither nonlinear probe becomes the new default: MLP2 yields BH-significant gains on CD4/CD8 AUROC (\(+0.0045\)) and CD4/CD8 balanced accuracy (\(+0.0049\)), but significantly worsens mono/macro balanced accuracy (\(-0.0076\)) and both stage endpoints. We therefore keep the linear 10D head as the main robust classifier result, and treat MLP2 as the best targeted nonlinear follow-up when CD4/CD8 discrimination is the priority.

\begin{figure}[t]
\centering
\includegraphics[width=\linewidth]{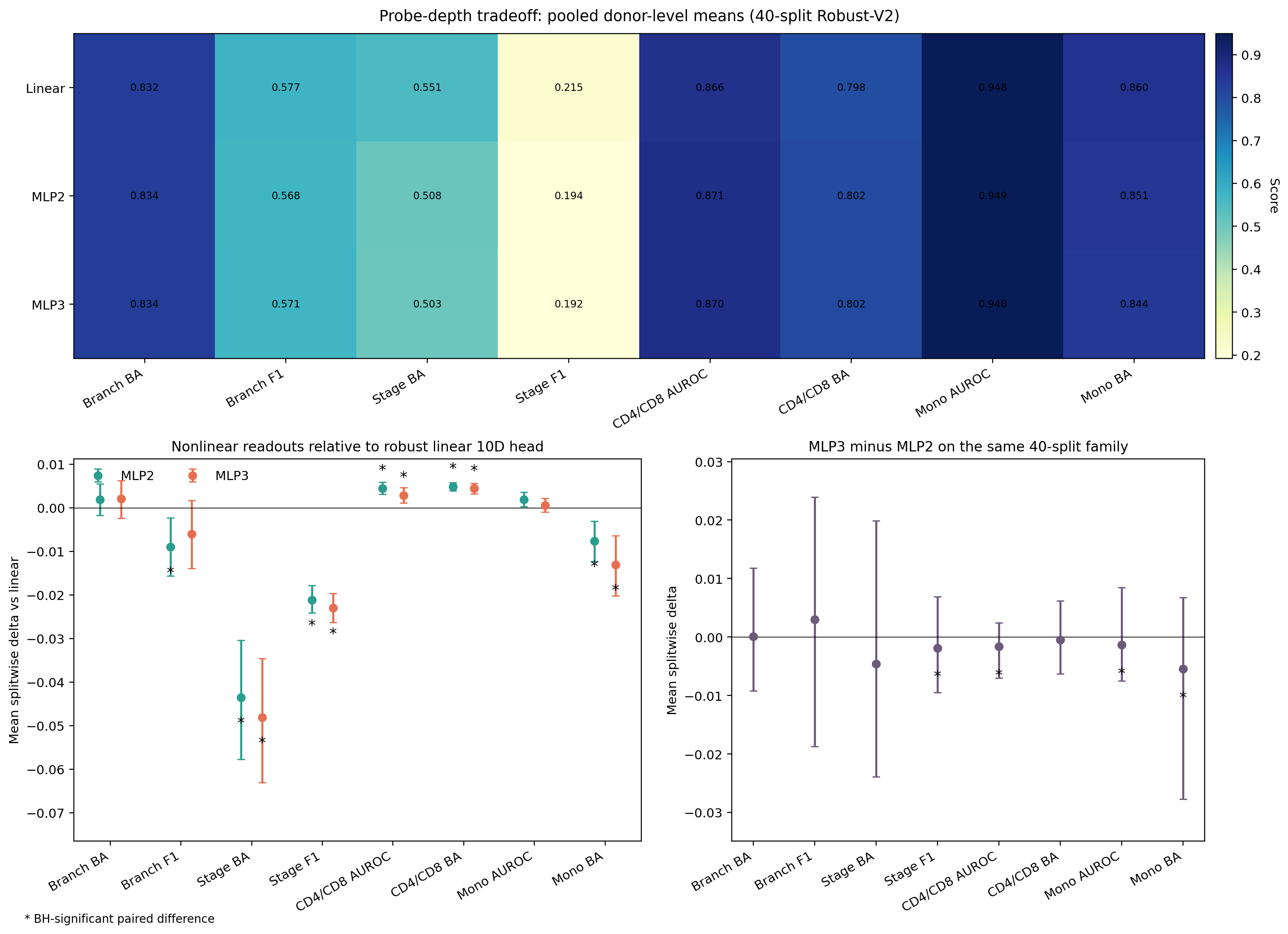}
\caption{Probe-depth tradeoff for the extracted 10D H65 cell head on the 40-split Robust-V2 family. The linear head remains the best overall default.}
\label{fig:head10d_probe_depth_tradeoff}
\end{figure}

\subsection*{S22. Direct learnability baseline}
To test whether a comparably small model trained directly on raw expression would recover the same task structure, we trained two direct raw-expression baselines on the same 12 grouped donor-holdout splits: a 1-hidden-layer model (\(1200\rightarrow64\rightarrow C\)) and a 2-hidden-layer model (\(1200\rightarrow128\rightarrow64\rightarrow C\)), both on \(\log(1+x)\) input.

\begin{table}[t]
\centering
\caption{Repeated grouped-split comparison between direct tiny-from-scratch raw-expression MLPs, frozen scGPT avg-pool + MLP, and the extracted 10D head + MLP2. Values are pooled donor-level means across 12 splits.}
\label{tab:direct_tiny_vs_extracted_metrics}
\small
\resizebox{\linewidth}{!}{
\begin{tabular}{lcccc}
\toprule
Metric & Frozen scGPT + MLP & Direct raw1200 \(1\times64\) & Direct raw1200 \(128\rightarrow64\) & Extracted 10D + MLP2 \\
\midrule
Branch balanced accuracy & 0.810 & 0.814 & 0.818 & \textbf{0.827} \\
Branch macro-F1 & 0.547 & 0.550 & \textbf{0.557} & 0.556 \\
Stage balanced accuracy & 0.403 & 0.499 & \textbf{0.504} & 0.468 \\
Stage macro-F1 & 0.157 & \textbf{0.190} & 0.186 & 0.180 \\
CD4/CD8 AUROC & 0.817 & 0.828 & 0.827 & \textbf{0.856} \\
CD4/CD8 balanced accuracy & 0.744 & 0.758 & 0.759 & \textbf{0.779} \\
Mono/Macro AUROC & 0.946 & 0.926 & 0.920 & \textbf{0.956} \\
Mono/Macro balanced accuracy & 0.844 & \textbf{0.866} & 0.863 & 0.865 \\
\bottomrule
\end{tabular}
}
\end{table}

Direct tiny models learn meaningful biology from raw expression, especially stage structure, but they do not recover the extracted head's stronger subtype-sensitive geometry. Neither direct model dominates the extracted head: relative to the extracted 10D+MLP2, both direct models are significantly worse on CD4/CD8 AUROC, CD4/CD8 balanced accuracy, and mono/macro AUROC. The extraction result is not reducible to ``just train a tiny MLP on raw input.''

\subsection*{S23. Runtime comparison}

\begin{table}[t]
\centering
\caption{Runtime decomposition for the repeated-split 4D-head versus frozen scGPT+MLP comparator.}
\label{tab:head4d_vs_frozen_scgpt_mlp_compute}
\small
\resizebox{\linewidth}{!}{
\begin{tabular}{lccc}
\toprule
Component & Scope & H65 cell head (4D) + linear & Frozen scGPT avg-pool + MLP \\
\midrule
Reference 1200-gene matrix extraction & one-time shared & \multicolumn{2}{c}{164.7 s} \\
Method-specific cache build & one-time & 99.5 s (drift + head training) & 5163.8 s (avg-pool extraction) \\
Per-split evaluation & mean / split & 8.8 s & 159.2 s \\
Trainable parameters / task & downstream & 5--170 & 172{,}610--174{,}690 \\
\bottomrule
\end{tabular}
}
\end{table}

The method-specific cache build is \(\sim99.5\) seconds for the extracted-head path versus \(\sim5163.8\) seconds for frozen scGPT avg-pool embeddings (\(\sim52\times\) gap). Per-split evaluation costs \(\sim8.8\) seconds for the head path versus \(\sim159.2\) seconds for the frozen-scGPT-MLP path (\(\sim18\times\) gap). Aggregated over the full 12-split campaign, the extracted-head path requires \(\sim204.8\) seconds versus \(\sim7074.4\) seconds for frozen scGPT (\(\sim34.5\times\) difference).

\subsection*{S24. T-cell resolution lens}
Permutation tests confirmed that T-cell subtype information (CD4 versus CD8) is genuinely present in the extracted features, not an artifact of dimensionality (CD4-vs-CD8 AUROC \(0.770\), permutation \(p=0.005\); naive-vs-mature AUROC \(0.752\), permutation \(p=0.005\)).

However, the default 3D visualization under-represents this signal: CD4-vs-CD8 AUROC is only \(0.646\) in the displayed 3D projection but \(0.822\) in the full 10D latent space, with the strongest single axis at dimension~4 (AUROC \(0.845\)). To address this, we introduced a regularized enhancement that amplifies the most discriminative T-cell axis while preserving the global manifold structure. At the selected regularization strength (\(\lambda=8\)), CD4-vs-CD8 AUROC improved from \(0.835\) to \(0.901\) with negligible loss in trustworthiness (\(\Delta=0.0018\)).

This motivated the \emph{T-cell separation lens}: a visualization tool that preserves the global manifold layout in the \(x/y\) plane while mapping the most discriminative T-cell axis to the \(z\)-coordinate. The lens improves interpretability of T-cell subtypes without changing the underlying quantitative metrics.

\begin{figure}[t]
    \centering
    \includegraphics[width=0.78\linewidth]{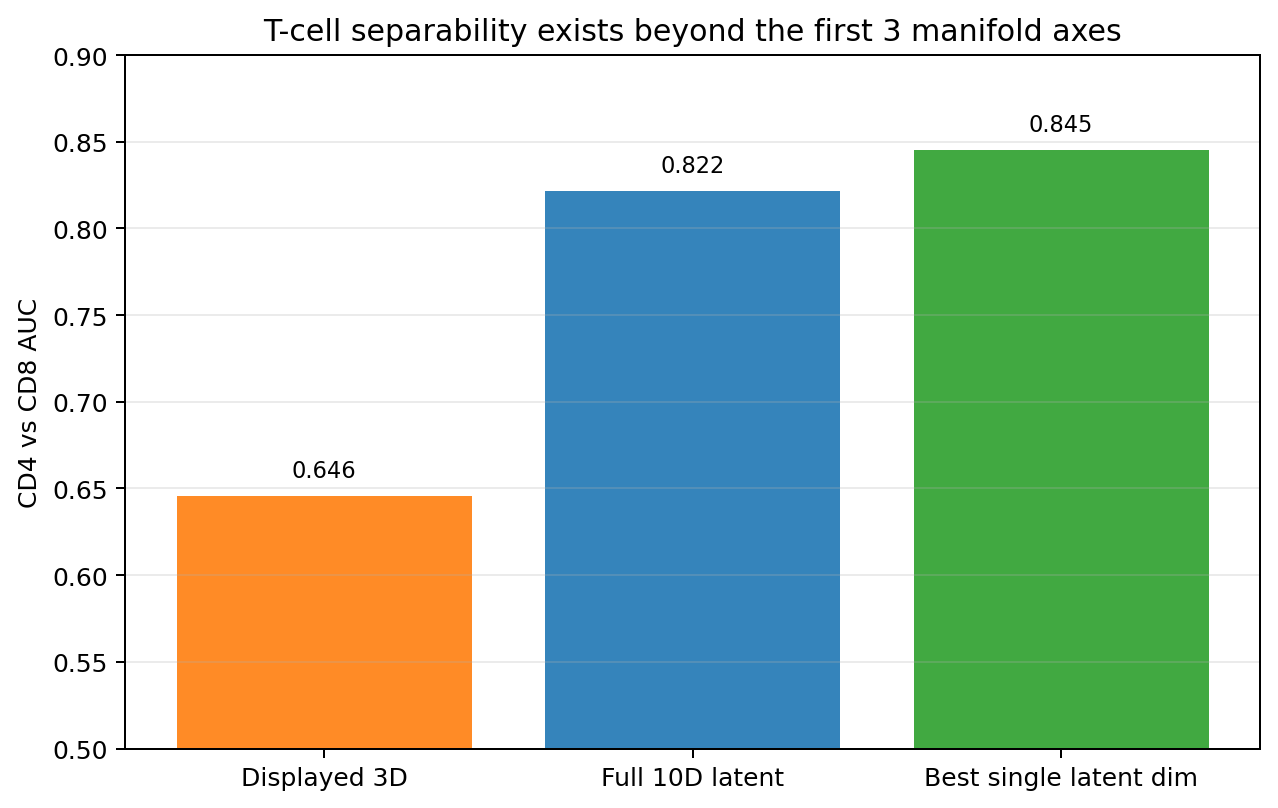}
    \caption{T-cell signal exists in latent space but is partially hidden in default 3D projection. The T-cell lens exposes this information for visualization.}
    \label{fig:tcell_auc}
\end{figure}

\textbf{Dimension-wise functional interpretation.} The first three latent dimensions are biologically meaningful but do not primarily encode CD4/CD8 separation:
\begin{itemize}[leftmargin=1.5em]
    \item \textbf{dim 1:} broad branch axis with strong B/plasma loading (AUC \(0.868\)); stage/branch eta\(^2\) \(0.895/0.869\).
    \item \textbf{dim 2:} strongest B/plasma contrast (AUC \(0.990\)) and moderate developmental-depth loading (\(|\rho|=0.381\) to HSC graph distance).
    \item \textbf{dim 3:} strongest erythroid trajectory signal among dims 1--3 (erythroid-path \(|\rho|=0.631\)), with moderate B/plasma loading (AUC \(0.854\)).
\end{itemize}
CD4/CD8 separation is concentrated outside the default first three axes, peaking at \textbf{dim 4} (AUC \(0.845\)); monocyte-vs-macrophage separation peaks at \textbf{dim 8} (AUC \(0.907\)). The strongest global stage/branch structure appears at \textbf{dim 9} (stage eta\(^2\) \(0.952\), branch eta\(^2\) \(0.883\)).

\subsection*{S25. Flatness and ripple structure}
\textbf{The manifold is predominantly flat in 3D but contains structured residual curvature.} When projected to 3D, the manifold is nearly planar: the best-fit 2D plane explains \(99.25\%\) of the 3D variance in the external map (\(97.17\%\) internally). However, the residuals are not random noise. After removing the dominant plane and any smooth quadratic trend, the remaining deviations show statistically significant periodic (ripple-like) structure (external \(R^2=0.171\), \(p=0.0033\); internal \(R^2=0.194\), \(p=0.0033\)).

\textbf{This structure persists in the full 10D latent space.} We tested each residual axis (after removing the two dominant principal components) for similar periodic structure. In the internal 10D space, all 8 tested residual axes showed significant periodicity; in the external space, 5 of 8 were significant after multiple-testing correction. The first two principal components captured most but not all variance (internal \(89.3\%\), external \(78.4\%\)), consistent with a compact but genuinely higher-dimensional manifold rather than a simple 2D surface.

\textbf{Calibration null check.} To verify that this test does not produce false positives on arbitrary high-dimensional point clouds, we generated a synthetic null dataset matched to the external 10D space in mean and covariance but with no biological structure. This null produced zero significant residual axes, confirming that the detected periodicity reflects real geometric structure in the manifold.

\subsection*{S26. Phase-2 deepening results}
\textbf{Latent interventions recover expected developmental directions.} On strict non-overlap external anchors, stem\(\rightarrow\)myeloid intervention showed strong monotonic enrichment (\(\rho=0.928\), \(p=4.9\times10^{-6}\), target fraction \(0.00\rightarrow1.00\)). Stem\(\rightarrow\)erythroid also showed a strong shift (\(\rho=0.739\), \(p=0.0039\), target fraction \(0.00\rightarrow0.727\)). CD4\(\rightarrow\)CD8 intervention was weaker and non-significant (\(\rho=-0.518\), \(p=0.070\)), consistent with subtype-local complexity.

\begin{figure}[t]
    \centering
    \includegraphics[width=0.88\linewidth]{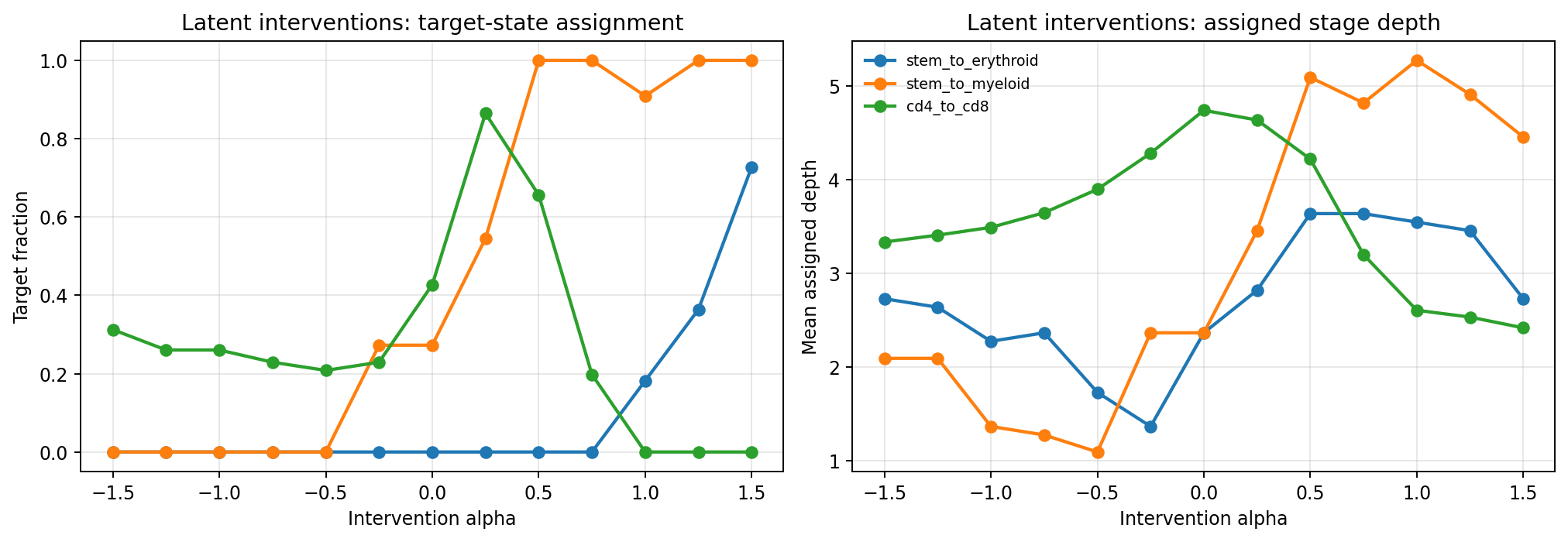}
    \caption{Phase-2 latent intervention sweeps. Positive monotonic shifts are strongest for stem\(\rightarrow\)myeloid and stem\(\rightarrow\)erythroid directions.}
    \label{fig:phase2_interventions}
\end{figure}

\textbf{Branch topology is preserved internally and partially externally.} In the internal manifold, the hematopoietic stem cell (HSC) region serves as the root node with connections to four major lineages, six branching points, and complete reachability to all terminal cell types. In the external non-overlap panel, the HSC root structure is preserved (three connections, five branching points) but reachability to terminal types is incomplete (\(57.1\%\)), likely reflecting the sparser external anchor coverage.

\textbf{Biological interpretation of the residual curvature remains open.} Despite the statistically significant periodic structure detected above, we could not assign biological meaning to specific residual axes. Associations between the periodic residual axes and a curated erythroid gene-program set did not survive multiple-testing correction (\(0/5\) significant at BH-FDR \(q\le0.05\)). Whether this residual geometry encodes additional lineage-specific information remains an open question for future work.

\subsection*{S27. Sparse factorization}
Starting from the selected \textbf{top1 + rank64} model, we applied a hard pruning grid over retained factor count and per-factor gene support size. The strongest sparse point (16 factors, 60 genes per factor, 10D with MLP2 readout) keeps 16 of the rank64 parent factors and retains the top 60 read genes plus top 60 write genes per factor. This reduces the deployable artifact to \(124{,}512\) bytes with only \(1{,}920\) active operator weights (\(\sim47\times\) smaller than dense compact top1). However, it no longer defines a new accuracy frontier: in the repeated 12-split audit it was materially worse than the rank64 parent on at least 6/8 BH-corrected endpoints.

The sparse-specific necessity audit shows that the top four retained parent factors \textbf{f01}, \textbf{f02}, \textbf{f00}, and \textbf{f03} explain \(68.9\%\) of the total sparse ablation impact, while the top eight explain \(87.2\%\). The same qualitative biology persists: \textbf{f01} is still the dominant branch factor, \textbf{f00} remains a strong stage factor, and \textbf{f03} remains the main mono/macrophage factor.

\begin{figure}[t]
    \centering
    \includegraphics[width=\linewidth]{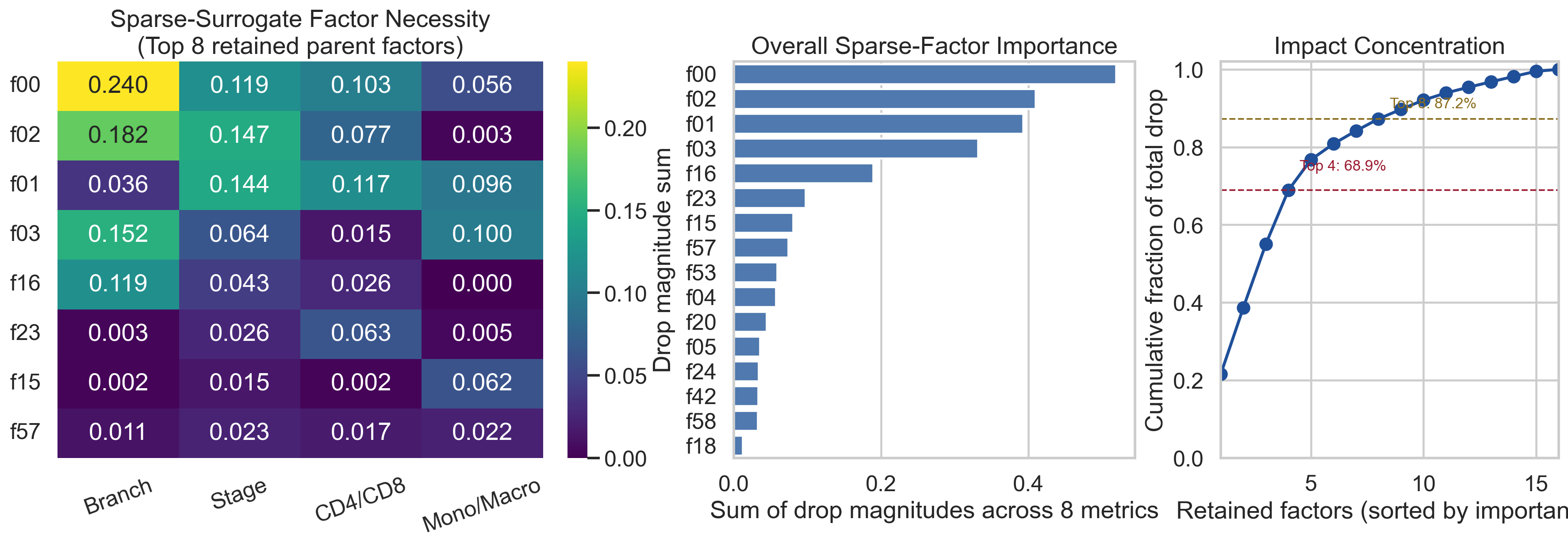}
    \caption{Fixed-probe leave-one-factor-out audit for the best sparse surrogate. The sparse model preserves the same mechanistic core as the rank64 parent: the top four retained factors explain \(68.9\%\) of total sparse ablation impact.}
    \label{fig:sparse_factor_ablation_summary}
\end{figure}

Local marker-set enrichment of the retained sparse read/write supports makes those factors biologically explicit: \textbf{f00} contrasts granulocytic/neutrophil programs against T/NK markers; \textbf{f01} aligns with monocyte/macrophage programs; \textbf{f02} contrasts lymphoid/B-cell-like structure against T/NK; and \textbf{f03} separates monocyte/macrophage-like programs from granulocytic structure.

\begin{table}[t]
\centering
\caption{Gene-program summary for the four core sparse-surrogate factors. Drop is the total pooled ablation impact. Read/write genes and enrichment are from cached local marker analysis. The same biological programs persist after hard-sparse pruning.}
\label{tab:sparse_factor_gene_programs}
\footnotesize
\resizebox{\linewidth}{!}{
\begin{tabular}{clclp{4.5cm}}
\toprule
Factor & Role & Drop & Top read/write genes & Enriched programs \\
\midrule
\textbf{f01} & Branch routing & 0.430 & R: EPB41, GBR1, VPEL3, GPX1, TAL001; W: DIF2R, EPB41, VPEL5, RBXP3L, RMT39 & Classical and non-classical monocyte identity \\[2pt]
\textbf{f02} & Lymphoid contrast & 0.254 & R: IL7R, GSTPL, DIF2R, ChrFTB, GPFC, FNL; W: HLA-DRA, MARCB9, AC96579.19, IGKC, IGLC1 & Na\"ive CD8$^+$ T-cell; thymic CD4$^+$; B-cell lineage \\[2pt]
\textbf{f00} & Stage ordering & 0.280 & R: IL7R, TENTSC, GSPT1, GVPC, GBK1, PRBX2; W: CSF3R, SLC2BA1, S100A8, ELR2, ACSL1, PRBB & Na\"ive CD8$^+$ T-cell; T-regulatory; granulopoiesis \\[2pt]
\textbf{f03} & Mono/macro struct. & 0.241 & R: LGALS3, BLC2L1, LV2, THBS1, C10A, BLVR8; W: VPEL1, ALPL, SLC28A3, FIG30R, ALT8, TENTSC & Macrophage; phagocyte; granulocyte/neutrophil \\
\bottomrule
\end{tabular}
}
\end{table}

A subsequent learned/data-aware sparse fit at the same scale (\(f=16\), \(k\in\{60,80\}\)) did \emph{not} rescue the sparse boundary. The better learned point remained clearly worse than the rank64 parent and also worse than the naive hard sparse point on branch balanced accuracy and CD4/CD8 AUROC. The interpretability value remains concentrated in the rank64 core and in the explicit hard-sparse boundary model.

\subsection*{S28. Rank64 versus sparse core comparison}
A direct comparison of the four core factors across the rank-64 and hard-sparse factorizations confirms that the same biological programs survive aggressive compression. Factors \textbf{f01}, \textbf{f02}, and \textbf{f03} retain identical top-ranked genes (both read and write sets) after sparsification, while \textbf{f00} preserves one gene set but changes the other, reflecting the information loss from pruning. Despite the smaller sparse model no longer matching the rank-64 parent on classification accuracy, both factorizations encode the same task-specialized pattern: \textbf{f01} dominates branch classification, \textbf{f02} separates lymphoid subtypes (B-cell versus T/NK), \textbf{f00} carries the strongest developmental-stage signal, and \textbf{f03} isolates monocyte/macrophage structure.

\subsection*{S29. Head/layer attribution details}
The 96-unit attribution scan showed that transferable hematopoiesis geometry is distributed across multiple layers, with clear high-performing units. The screen-selected best unit was \textbf{Layer 2 / Head 5} (L2H5). After full refit, L2H5 reached external corr \(0.636\), residualized corr \(0.634\), and trustworthiness \(0.985\), with 3D subtype AUCs \(0.711\) (CD4/CD8) and \(0.873\) (monocyte/macrophage).

\begin{figure}[t]
    \centering
    \includegraphics[width=0.82\linewidth]{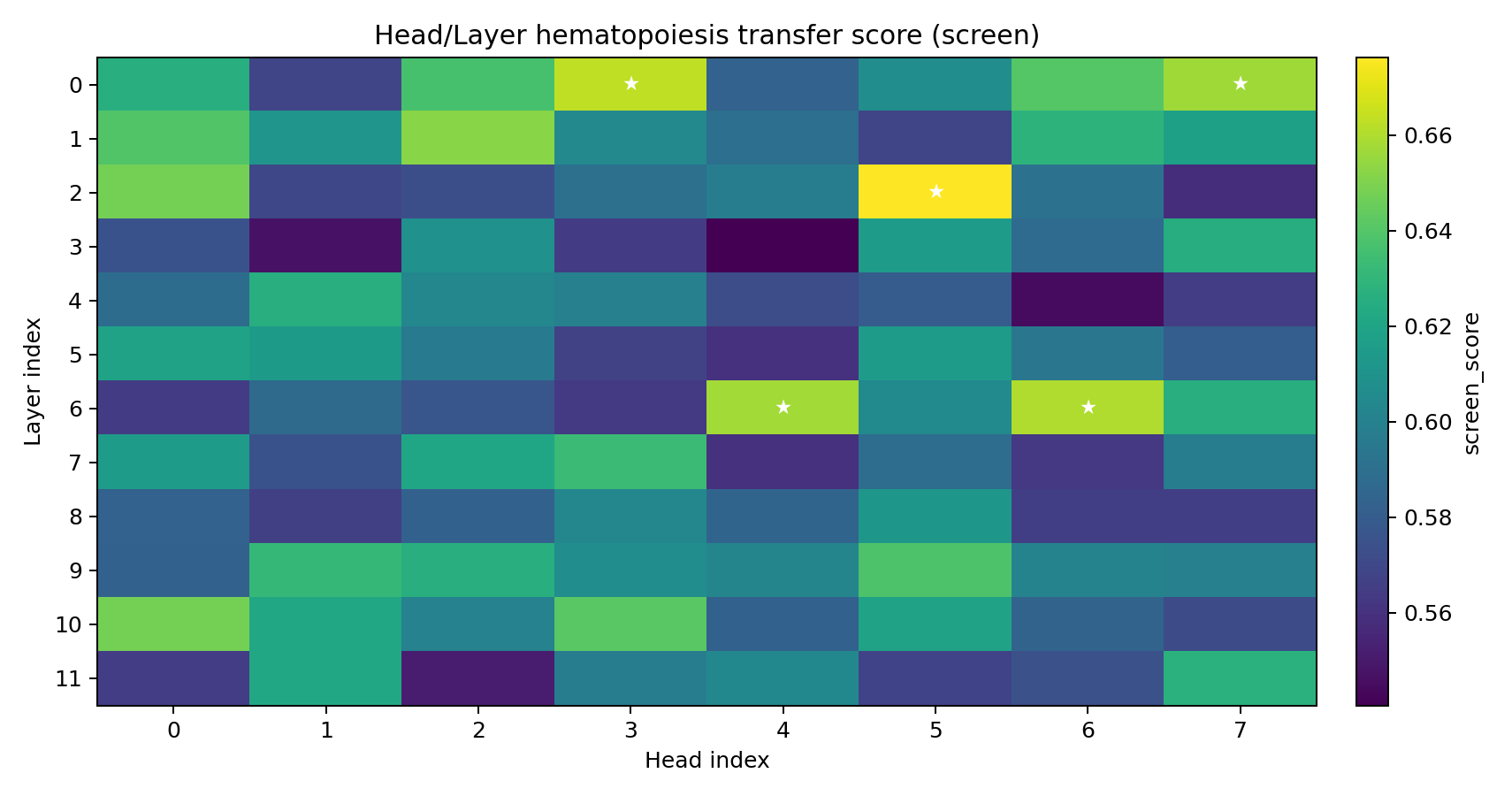}
    \caption{Head/layer attribution screen across all 96 scGPT attention units. Color shows external transfer screen score; stars mark top-ranked units.}
    \label{fig:head_attr_heatmap}
\end{figure}

\begin{table}[t]
\centering
\caption{Top single heads after full refit on internal anchors and zero-shot evaluation on strict non-overlap external anchors.}
\label{tab:head_attr_top5}
\small
\begin{tabular}{lcccc}
\toprule
Unit & Corr\(_{\mathrm{resid}}\) & Trustworthiness & CD4/CD8 AUC\(_{3D}\) & Mono/Macro AUC\(_{3D}\) \\
\midrule
L2H5 (screen rank 1) & 0.634 & 0.985 & 0.711 & 0.873 \\
L0H3 (screen rank 2) & 0.627 & 0.981 & 0.743 & 0.769 \\
L6H6 (screen rank 3) & 0.621 & 0.984 & 0.765 & 0.867 \\
L6H4 (screen rank 4) & 0.610 & 0.986 & 0.765 & 0.810 \\
L0H7 (screen rank 5) & 0.636 & 0.977 & 0.746 & 0.765 \\
\bottomrule
\end{tabular}
\end{table}

Relative to pooled transferable heads, the best single head is competitive for global stage-distance alignment (L2H5 residualized corr \(0.634\) vs pooled-anchor \(0.605\), pooled-cell \(0.612\), and conservative-hybrid \(0.632\)). However, subtype separation remains better in pooled/hybrid variants (CD4/CD8 AUC \(0.911\) in conservative hybrid vs \(0.711\) in L2H5), supporting a two-view interpretation: single heads can carry strong global geometry, while pooled objectives can better expose fine subtype margins.

\subsection*{S30. Factor ablation details (core-sufficiency and interactions)}
Necessity does not imply sufficiency. We ran a fixed-probe core-sufficiency test on cumulative subsets \(\{\textbf{f01}\}\), \(\{\textbf{f01},\textbf{f02}\}\), \(\{\textbf{f01},\textbf{f02},\textbf{f00}\}\), and \(\{\textbf{f01},\textbf{f02},\textbf{f00},\textbf{f03}\}\) over the same 12 grouped donor-holdout splits. The four-factor core is the strongest of these subsets, but it still falls well short of the intact rank64 model: pooled means are branch balanced accuracy \(0.572\) versus \(0.820\), stage balanced accuracy \(0.191\) versus \(0.435\), CD4/CD8 AUROC \(0.699\) versus \(0.846\), and mono/macro AUROC \(0.834\) versus \(0.951\), with BH-significant losses on all 8 endpoints.

The sweep is clearly non-additive: adding \textbf{f00} before \textbf{f03} sharply damages mono/macro performance (mono/macro AUROC \(0.381\) for \(\{\textbf{f01},\textbf{f02},\textbf{f00}\}\)), and only partial recovery appears once \textbf{f03} is restored.

The exhaustive 15-subset interaction sweep resolves this further. No pair or triple closes the gap to the intact rank64 model: all 15 subsets remain BH-significantly worse on all 8 endpoints. But different subsets dominate different tasks. Branch and stage are best recovered by the full four-factor set; mono/macro is best recovered by the pair \(\{\textbf{f01},\textbf{f03}\}\) (mono/macro AUROC \(0.902\), balanced accuracy \(0.803\)); and the strongest CD4/CD8 recovery comes from triples centered on \(\textbf{f00}\) and \(\textbf{f03}\) (CD4/CD8 AUROC \(0.703\)).

\begin{table}[t]
    \centering
    \caption{Best subset from the exhaustive 15-subset interaction sweep for each audited endpoint. Every listed subset remains BH-significantly below the intact rank64 model.}
    \label{tab:rank64_core_interaction_best_subsets}
    \small
    \begin{tabular}{@{}llll@{}}
        \toprule
        Endpoint & Best subset & Size & Best / intact \\
        \midrule
        Branch BA & \(\{\textbf{f01},\textbf{f02},\textbf{f00},\textbf{f03}\}\) & 4 & 69.7\% \\
        Branch F1 & \(\{\textbf{f01},\textbf{f02},\textbf{f00},\textbf{f03}\}\) & 4 & 71.7\% \\
        Stage BA & \(\{\textbf{f01},\textbf{f02},\textbf{f00},\textbf{f03}\}\) & 4 & 44.0\% \\
        Stage F1 & \(\{\textbf{f01},\textbf{f02},\textbf{f00},\textbf{f03}\}\) & 4 & 45.5\% \\
        CD4/CD8 AUROC & \(\{\textbf{f02},\textbf{f00},\textbf{f03}\}\) & 3 & 83.1\% \\
        CD4/CD8 BA & \(\{\textbf{f01},\textbf{f00},\textbf{f03}\}\) & 3 & 82.9\% \\
        Mono/Macro AUROC & \(\{\textbf{f01},\textbf{f03}\}\) & 2 & 94.9\% \\
        Mono/Macro BA & \(\{\textbf{f01},\textbf{f03}\}\) & 2 & 92.3\% \\
        \bottomrule
    \end{tabular}
\end{table}

\subsection*{S31. Dimension-wise functional audit}
\textbf{Cell-trained head dimension audit.} When training the adaptor head on individual cells rather than anchors, the overall stage and branch information remains strong (best stage \(\eta^2\): dim~3, \(0.924\); best branch \(\eta^2\): dim~8, \(0.835\)), but the distribution of subtype information across dimensions changes substantially. CD4/CD8 separation now peaks at \textbf{dim~10} (AUROC \(0.902\)) rather than dim~4 (\(0.845\) in the anchor-trained head). The first three dimensions---which determine the default 3D visualization---carry much more T-cell subtype contrast in the cell-trained head (max CD4/CD8 AUROC among dims~1--3: \(0.847\) vs \(0.607\); max mono/macro AUROC: \(0.936\) vs \(0.659\)), at the cost of reduced B/plasma dominance in those early dimensions (\(0.910\) vs \(0.990\)).

\textbf{Temporal differentiation is distributed, not single-axis.} In the anchor-trained head, the strongest global depth proxy loads on dim 7 (\(\rho=0.452\) to HSC graph distance), while lineage-specific ordering peaks on different axes (erythroid on dim 10, \(|\rho|=0.672\); monocyte/macrophage on dim 8, \(|\rho|=0.731\); B\(\rightarrow\)plasma on dim 2, \(|\rho|=0.875\)). In the cell-trained head, depth loading shifts to dim 9 (\(\rho=0.640\)), again with lineage-specific maxima on different dimensions.

\textbf{Composite temporal axis search.} A ridge-regularized linear axis over LET-10D using six standardized temporal components reached Spearman \(\rho=0.623\) to the composite target and \(\rho=0.627\) to HSC depth (\(p=5.0\times10^{-4}\), one-sided permutation), with dominant loadings on dims 9, 2, 5, and 4.

\textbf{Important caveat: dimensions are not independent factors.} The latent axes are correlated in both variants (mean absolute off-diagonal correlation: anchor-trained \(0.493\), cell-trained \(0.400\)). Interpretation should treat dimensions as a mixed/distributed code rather than one-factor-per-axis.

\subsection*{S32. H38 goal and biological target}
To test whether the discovery protocol generalizes beyond hematopoiesis, we ran an independent branch (H38) targeting \emph{intercellular communication geometry} in scGPT. The target ruler differed from H65: instead of developmental-stage distances, H38 used curated interaction-path structure derived from OmniPath \cite{turei2021omnipath} and ligand-receptor attention-flow summaries. The question was whether a compact manifold fit to this signal could pass strict baseline and transfer checks under the same no-cheating policy used in the main study.

\subsection*{S33. H38 canonical setup}
H38 used an internal anchor panel of 104 anchors constructed from 30 selected ligand--receptor interaction-path pairs (square-root-weighted distances). We evaluated three random seeds, two manifold-learning methods (geodesic MDS and Isomap), and three neighborhood sizes (\(k\in\{10,15,20\}\)). The best internal setting was seed~43 with geodesic MDS at \(k=10\).

Median internal summary across settings:
\begin{itemize}[leftmargin=1.5em]
    \item trustworthiness \(=0.9082\) (min/max \(0.8910/0.9282\)),
    \item geodesic-anchor correlation \(=0.8277\) (min/max \(0.8256/0.8297\)),
    \item blocked-permutation \(p=0.0005\),
    \item manifold-minus-PCA2 baseline margin \(=+0.0544\).
\end{itemize}

The canonical internal run passed both primary and strict baseline gates, with stable fragility controls (subsample sign consistency \(=1.0\), cell-type-holdout sign consistency \(=1.0\)).

\begin{figure}[ht]
    \centering
    \includegraphics[width=0.78\linewidth]{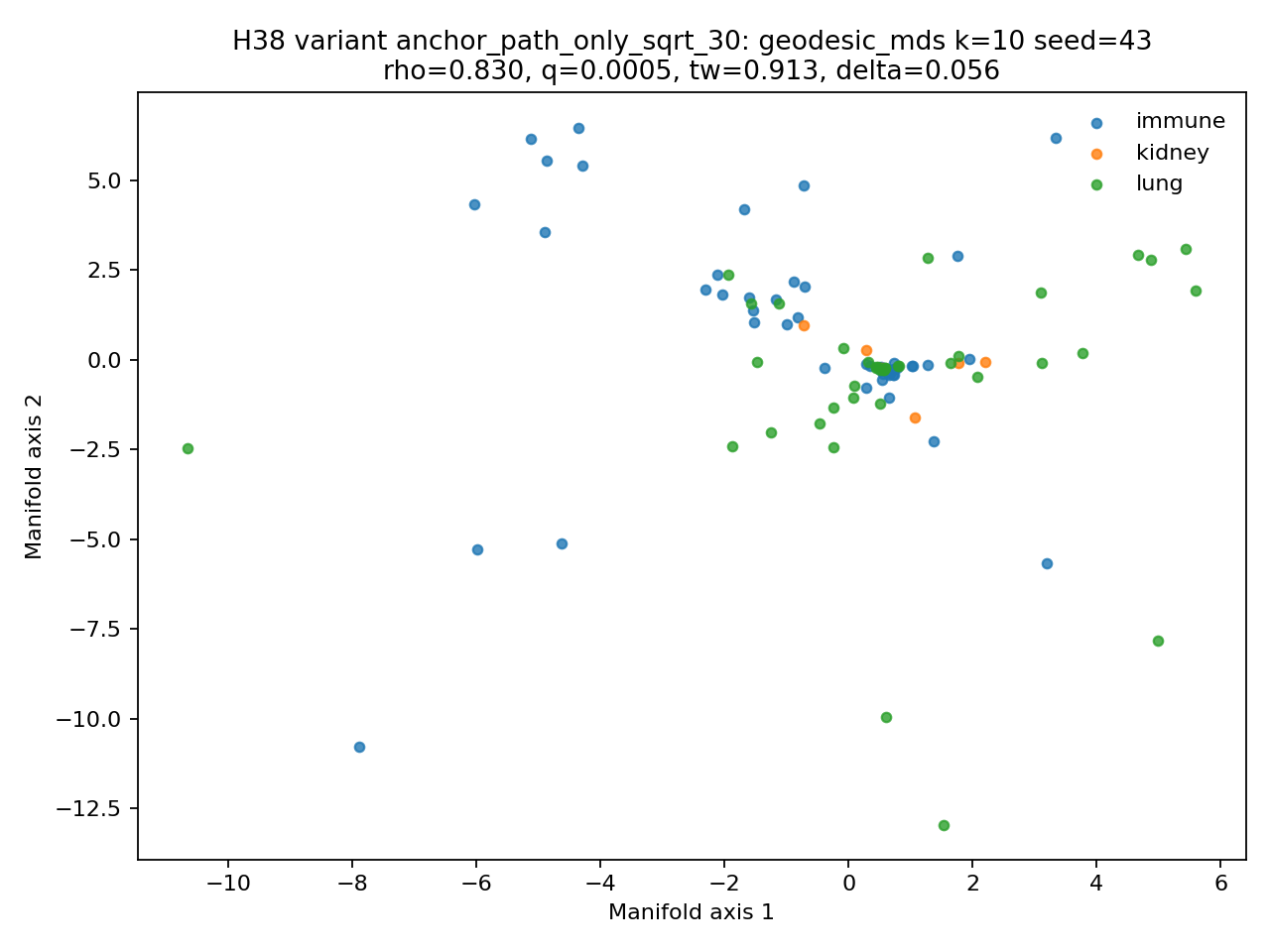}
    \caption{\textbf{H38 internal canonical manifold.} Best internal embedding using 30 interaction-path pairs with square-root-weighted distances. This branch captures a structured intercellular-communication axis system distinct from the hematopoietic developmental geometry in the main text.}
    \label{fig:supp_h38_internal}
\end{figure}

\subsection*{S34. H38 first external check: strong association but strict-baseline failure}
The first zero-shot external replication used a preprocessed immune panel (20,000 cells, 102 tissue-level anchors after requiring at least 40 cells per anchor group). It remained strongly significant by correlation and permutation test, but failed strict baseline gates:
\begin{itemize}[leftmargin=1.5em]
    \item trustworthiness median \(=0.9373\),
    \item correlation median \(=0.9086\),
    \item blocked-permutation \(p=0.0005\),
    \item median margin vs PCA2 \(=-0.0736\) (non-passing under strict baseline policy).
\end{itemize}
This reproduced the same pattern observed earlier in our broader loop: high raw association alone is insufficient for a positive claim unless baseline margins also pass.


\subsection*{S35. H38 negative controls and axis-level interpretation}
Negative controls showed that not all null manipulations were cleanly separated from the observed score under this first external setup, especially for strict baseline margins:
\begin{itemize}[leftmargin=1.5em]
    \item flow-gene permutation: empirical \(p_{corr}=0.0732\), \(p_{\Delta}=0.2439\),
    \item random hop-matched pairs: empirical \(p_{corr}=0.3659\), \(p_{\Delta}=0.9756\).
\end{itemize}
Axis interpretation remained biologically meaningful, with strongest module associations in cytokine/chemokine, IL1-axis, coagulation/ECM, and lipid/metabolic programs (module-pair counts \(8,4,17,2\), respectively). For example, axis 1 showed strong broad-class structure (\(\eta^2=0.735\)) and opposite-direction associations with cytokine/chemokine (\(\rho=-0.549\)) and lipid/metabolic (\(\rho=0.566\)).

\begin{figure}[ht]
    \centering
    \includegraphics[width=0.72\linewidth]{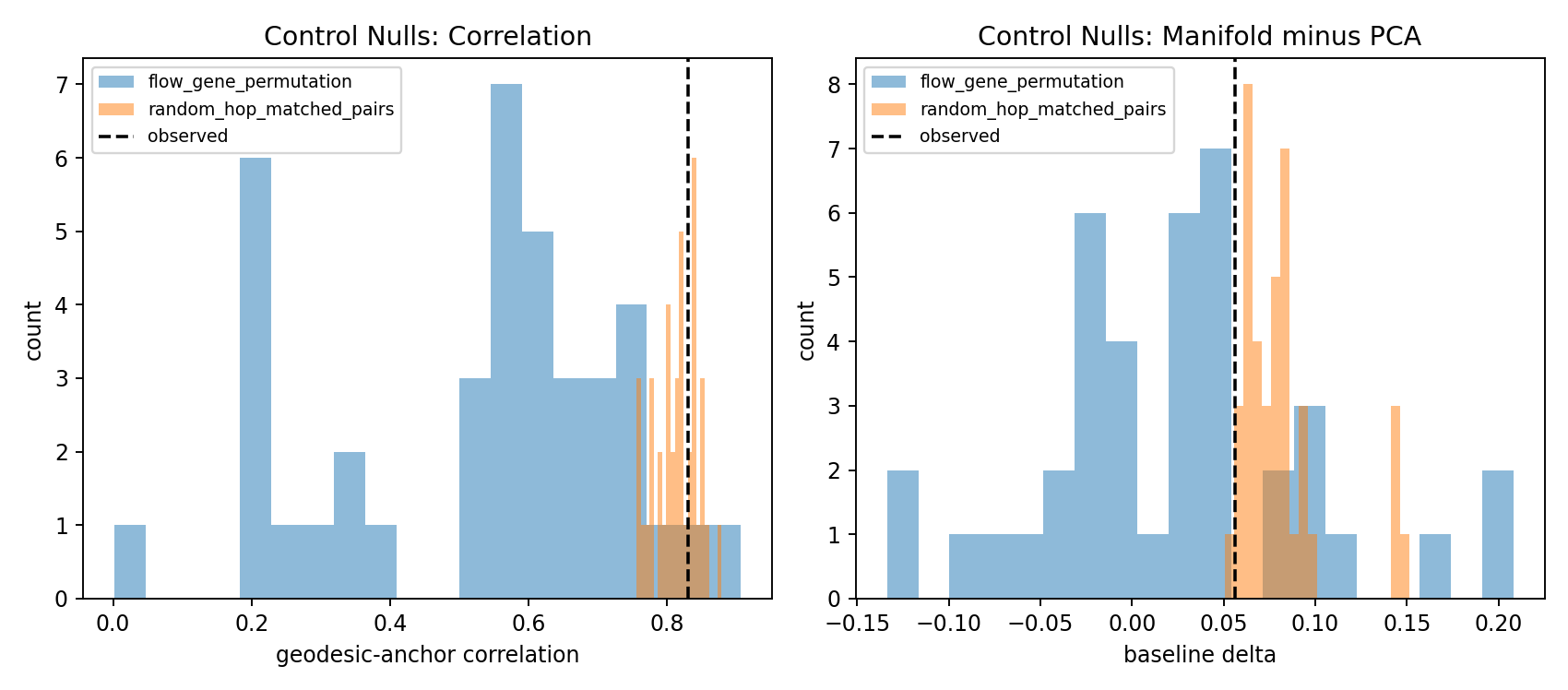}
    \caption{\textbf{H38 negative-control distributions.} Observed H38 alignment was not cleanly separated from all control draws under early external conditions, motivating the rescue program below.}
    \label{fig:supp_h38_controls}
\end{figure}

\subsection*{S36. H38 iterative rescue without target-dataset leakage}
We ran an iterative rescue track (v4\(\rightarrow\)v11), keeping external evaluation frozen and avoiding any target-dataset optimization. Table~\ref{tab:supp_h38_rescue} summarizes the progression.

\begin{table}[ht]
\centering
\caption{H38 rescue progression on external zero-shot evaluation (minimum 40 cells per anchor group).}
\label{tab:supp_h38_rescue}
\small
\begin{tabular}{lcc}
\toprule
Run (key change) & External corr & \(\Delta\) vs PCA2 \\
\midrule
v4 (pair hygiene + hard negatives) & 0.8850 & -0.0909 \\
v5 (cell-trained transfer head) & 0.9624 & -0.0199 \\
v7 (metric head + cleaned pair set + ensemble) & 0.9707 & -0.0061 \\
v9 (cell-metric head + ensemble) & 0.9800 & +0.0032 \\
v10 (MLP-geom/dim6 extras) & 0.9747 & -0.0021 \\
v11 (raw diverse 100k-cell internal training + ensemble) & 0.9825 & +0.0057 \\
\bottomrule
\end{tabular}
\end{table}

The final v11 run expanded internal training coverage from narrow subsets to raw Tabula immune+lung+kidney anchors:
\begin{itemize}[leftmargin=1.5em]
    \item internal anchors: 392,
    \item sampled training cells: 100,000 (target met exactly),
    \item diversity: 23 donors, 58 cell types, 3 tissues.
\end{itemize}

On the external panel (minimum 40 cells per anchor), v11 produced:
\begin{itemize}[leftmargin=1.5em]
    \item anchor baseline: corr \(0.8862\), \(\Delta\) vs PCA2 \(=-0.0906\),
    \item cell-metric head: corr \(0.9776\), \(\Delta\) vs PCA2 \(=+0.0008\),
    \item ensemble head: corr \(0.9825\), \(\Delta\) vs PCA2 \(=+0.0057\),
    \item blocked-permutation \(p=0.0005\).
\end{itemize}
This closes the main external-validation gap for H38 under the same strict criteria used in the main paper: no target-dataset information was used during any rescue iteration.

\subsection*{S37. H38 practical interpretation and limits}
H38 supports a complementary biological signal family to hematopoiesis. While H65 organizes developmental branch structure, H38 emphasizes communication-program geometry (cytokine/chemokine- and IL1-linked axes with ECM/coagulation and lipid/metabolic contrasts). The rescue sequence also clarifies a methodological point: transfer quality depended strongly on internal training coverage and diversity, not on loosening gates.

At the same time, H38 remains secondary evidence relative to H65:
\begin{itemize}[leftmargin=1.5em]
    \item it required a longer rescue path to pass strict external margins,
    \item the biological interpretation of its axes is suggestive but has not been validated with the same intervention experiments used for H65,
    \item classification benchmark results are mixed (for example, the 2D manifold embedding outperformed a 2D PCA baseline but not a 6D PCA baseline on grouped macro-F1: \(0.482\) vs \(0.464\) vs \(0.531\)).
\end{itemize}

We therefore treat H38 as a successful generalization case: the discovery and extraction protocol transferred to a biologically distinct manifold family (intercellular communication rather than developmental hierarchy) and achieved strict external validation after expanding internal training coverage, but with lower overall maturity than the primary hematopoiesis result.

\end{document}